\documentclass{article} 
\usepackage{iclr2024_conference,times}

\usepackage{hyperref}
\usepackage{url}

\usepackage{tabularray}
\usepackage[utf8]{inputenc} 
\usepackage[T1]{fontenc}    
\usepackage{hyperref}       
\usepackage{url}            
\usepackage{booktabs}       
\usepackage{amsfonts}       
\usepackage{nicefrac}       
\usepackage{microtype}      
\usepackage{graphicx}
\usepackage{amsmath,amssymb,amsfonts,mathtools,bm}
\usepackage{caption}
\usepackage{multirow}
\usepackage{array}
\usepackage{algorithm, algorithmic}

\usepackage{arydshln}
\usepackage{hyperref}

\newcommand{\bracket}[3]{\left#1 #3 \right#2}
\newcommand{\mbracket}[5]{\left#1 #4 \middle#2 #5 \right#3}
\renewcommand{\b}{\bracket{(}{)}}
\newcommand{\bc}{\mbracket{(}{\vert}{)}}

\newcommand{\bareP}{\operatorname{P}}
\renewcommand{\P}[1][]{\bareP_{#1}\b}
\newcommand{\Pc}[1][]{\bareP_{#1}\bc}
\newcommand{\param}{{\boldsymbol{\theta}}}
\newcommand{\paramMAP}{{\param_\text{MAP}}}
\newcommand{\data}{\mathcal{D}}
\renewcommand{\L}[1]{\mathcal{L}(\y, \X; #1)}
\newcommand{\N}{\mathcal{N}\b}
\renewcommand{\S}{\mathbf{\Sigma}}
\newcommand{\G}{\mathbf{G}}
\newcommand{\I}{\mathbf{I}}
\newcommand{\J}{\mathbf{J}}
\newcommand{\y}{\mathbf{y}}
\newcommand{\X}{\mathbf{X}}
\newcommand{\Lm}{\mathbf{L}}
\newcommand{\W}{\mathbf{W}}
\newcommand{\A}{\mathbf{A}}
\newcommand{\F}{\mathbf{F}}
\renewcommand{\a}{\mathbf{a}}
\newcommand{\g}{\mathbf{g}}
\newcommand{\B}{\mathbf{B}}
\newcommand{\x}{\mathbf{x}}
\newcommand{\mL}{\mathbf{\Lambda}}
\newcommand{\softmax}{\operatorname{softmax}}
\newcommand{\diag}{\operatorname{diag}}
\newcommand{\argmax}{\operatorname*{argmax}}

\newcommand{\nin}{n_\text{in}}
\newcommand{\nout}{n_\text{out}}
\newcommand{\nlr}{n_\text{lr}}

\hypersetup{
    colorlinks,
    linkcolor={red!50!black},
    citecolor={blue!50!black},
    urlcolor={blue!80!black}
}

\title{Bayesian low-rank adaptation for large language models}

\author{%
Adam X. Yang$^{1*}$ \quad Maxime Robeyns$^{1*}$ \quad Xi Wang$^{2}$ \quad
Laurence Aitchison$^1$ \\
$^1$University of Bristol \quad $^2$University of Massachusetts, Amherst \\
\texttt{\{adam.yang,maxime.robeyns.2018,laurence.aitchison\}@bristol.ac.uk}\\
\texttt{xwang3@cs.umass.edu}\\
}

\iclrfinalcopy 
\begin{document}

\maketitle

\begin{abstract}
Low-rank adaptation (LoRA) has emerged as a new paradigm for cost-efficient fine-tuning of large language models (LLMs). However, fine-tuned LLMs often become overconfident especially when fine-tuned on small datasets. Bayesian methods, with their inherent ability to estimate uncertainty, serve as potent tools to mitigate overconfidence and enhance calibration. In this work, we introduce Laplace-LoRA, which applies a Bayesian approach to the LoRA parameters. Specifically, Laplace-LoRA applies a Laplace approximation to the posterior over the LoRA parameters, considerably improving the calibration of fine-tuned LLMs.
\end{abstract}

\section{Introduction}
In recent years, fine-tuning large language models (LLMs) have become increasingly important \citep{houlsby2019parameter,hu2021lora,liu2022few,ding2022delta,opendelta}. 
Fine-tuning is used both to adapt LLMs for specific tasks and to create general instruction-following models (e.g.\ using Reinforcement Learning from Human Feedback; RLHF \citealp{wei2021finetuned,ouyang2022training,chung2022scaling,wang2022self}).

However, fine-tuned LLMs have a notable limitation: they often exhibit overconfidence \citep{jiang2021can,xiao2022uncertainty,he2023preserving,tian2023just,openai2023gpt4}.
This is particularly problematic in safety-critical applications or when making decisions in areas where limited data is available, such as medical diagnosis, finance and experimental design \citep{singhalLargeLanguageModels2022, wuBloombergGPTLargeLanguage2023, lampinenPassiveLearningActive2023, li2022pre}.
Consequently, there is an urgent need for strategies that enhance the calibration of fine-tuned LLMs, ensuring that their predictions are as trustworthy as they are powerful.

Bayesian deep learning is commonly proposed as a solution to overconfidence in deep networks \citep[e.g.][]{blundell2015weight,zhang2019cyclical,kristiadi2020being,ober2021global,fortuin2021bayesian,aitchison2021deep}.
Historically, the field of Bayesian deep learning has frequently considered ResNets for image classification \citep{shridhar2019comprehensive,dusenberry2020efficient,izmailov2021bayesian}.
While there are a few previous papers on Bayesian language models \citep[e.g.][]{tran2019bayesian,xue2021bayesian,fan2020bayesian,cinquin2021pathologies,zhang2019cyclical,chen2023calibrating}, most of them are focused on language model pre-training (though see Related Work for exceptions).
These methods have not yet seen widespread adoption, potentially because they are costly to scale up to larger models, and large-scale pre-trained models are already reasonably well-calibrated \citep{openai2023gpt4}.

Importantly, the advantages of Bayesian fine-tuning are much clearer than the advantages of Bayesian pretraining.
First, fine-tuned models are typically poorly calibrated \citep{jiang2021can,xiao2022uncertainty,he2023preserving,tian2023just}, even after very large-scale instruction fine-tuning \citep{openai2023gpt4}, and Bayes might help us improve this calibration.
Second, this poor calibration may arise in fine-tuning settings, as there is often far less data available in fine-tuning than pre-training. 
Bayes might help us to reason about uncertainty given these limited datasets.
Finally, the computational burden of fine-tuning is typically much smaller than that of pre-training, implying more headroom for any additional computational costs that might arise as a result of Bayesian inference.

However, LLMs can be very large. 
Indeed, fine-tuning all the weights of an LLM is often prohibitively costly, in which case Bayesian fine-tuning will of course also not be possible.
Instead, a number of more efficient approaches have emerged under the banner of parameter-efficient fine-tuning (PEFT \citealp{liu2022few,ding2022delta,opendelta,shi2023dept}).
PEFT methods tend to train a small number of additional parameters on top of the underlying fixed-pre-trained LLM. 
For instance, a common approach is to fine-tune only low-rank adapters for each weight matrix (LoRA; \citep{hu2021lora}).
PEFT approaches such as LoRA have democratized LLM fine-tuning and are implemented in a number of widely adopted libraries including OpenDelta \citep{opendelta} and PEFT \citep{peft}.

Importantly, PEFT approaches such as LoRA should also enable more efficient Bayesian finetuning.
We develop the first Bayesian inference method designed specifically for the LoRA parameters in LLM fine-tuning.
Specifically, we use a post-hoc Laplace approximation \citep{mackay1992practical,ritter2018scalable,daxberger2021bayesian,daxberger2021laplace,antoran2022adapting}, which allows us to keep the standard, highly efficient pre-training and fine-tuning pipelines \textit{exactly} the same.
We then use Laplace to compute the uncertainties over just the small number of LoRA \citep{hu2021lora} parameters.
We call the resulting method Laplace-LoRA, and we show dramatic improvements in calibration on fine-tuned LlaMA2-7B on six common sense reasoning tasks and in out-of-distribution settings.

\section{Related work}
Past work on integrating Bayesian inference with language models has usually operated in the large-scale pre-training setting \citep{tran2019bayesian,xue2021bayesian,cinquin2021pathologies,zhang2019cyclical,chen2023calibrating}, where the advantages of Bayes are unclear, because pre-training datasets are very large and large-scale pretrained models seem to be reasonably well-calibrated even without Bayes \citep{kadavath2022language,openai2023gpt4}.
In contrast, our work operates in the fine-tuning setting, where the advantages of Bayes are far more evident, e.g. because even large-scale instruction fine-tuning gives poor calibration \citep{openai2023gpt4}.

While \citet{fan2020bayesian} and \citet{zhang2021bayesian} do consider fine-tuning, they make the unusual choice to define a prior and approximate posterior over the attention weights, rather than over parameters (i.e.\ weight matrices).
Our method --- Laplace LoRA --- has two key benefits over their approach.
First, Laplace LoRA is a post-hoc method that keeps the pretraining and fine-tuning process \textit{exactly} the same. Therefore Laplace LoRA is able to exploit very efficient pre-existing implementations of these methods \citep{ding2022delta,peft,dettmers2023qlora}, which is not possible when defining priors and approximate posteriors over attention weights.
Second, Laplace LoRA massively reduces the dimensionality of Bayesian inference problem, either as against full finetuning where we would need to reason about the posterior over all weights, or as against \citet{fan2020bayesian} and \citet{zhang2021bayesian}, which must reason about the posterior over attention weights at every layer for every fine-tuning datapoint. 
There are 32 attention heads per layer and 32 layers in LLaMA 7B, if we assume 1000 sequences of length 30 in the finetuning set, that gives $32$ layers $\times 32$ heads per layer $\times 30^2$ attention weights per head $\times$ 30 sequences $\approx$ one billion attention weights. 
That one billion attention weights contrasts to only around 6 million LoRA parameters in our finetuning.

Laplace inference for Bayesian neural networks is well-studied \citep{ritter2018scalable,kristiadi2020being,immer2021improving,antoran2022adapting,daxberger2021laplace,deng2022accelerated}.
However, the only application of Laplace approximations to language models that we know of comes from \citet{daxberger2021laplace} which used DistillBERT (a 66 million parameter model) as a feature-extractor, and applied Laplace inference only at the final linear last-layer.  
In contrast, in our work we consider Laplace inference at all layers in models 100 times larger, by using LoRA adapters.

A separate line of research has been dedicated to regularizing language model fine-tuning to improve calibration. 
For instance, Mixout \citep{lee2019mixout} stochastically substitutes model weights with their pre-trained counterparts.
\citet{park2022calibration} adapted Mixup \citep{zhang2017mixup} to augment the dataset during language model fine-tuning. Meanwhile, \citet{he2023preserving} introduced a KL regularization between the output distributions of both fine-tuned and pre-trained language models, and added an $L_2$ regularization to the final embedding vector. However, their regularization relies on the masked language model objective which only applies to BERT-like models.
Critically, this body of work is orthogonal to our work in that they give better settings for the fine-tuned weights.
Laplace-LoRA does not change the fine-tuned weights at all, so it could use weights from standard fine-tuning, or it could use the improved weights given by these methods.
Instead, Laplace-LoRA uses a Laplace approximation to estimate uncertainty around any given mean value of the weights.
Laplace-LoRA thus has the potential to improve calibration for \textit{any} good method for generating fine-tuned weights.


\section{Background}

\subsection{Low-rank adaptation (LoRA)}

LLMs have a large number of large weight matrices, denoted $\W_0 \in \mathbb{R}^{\nout\times\nin}$, with inputs $\a$ and outputs $\mathbf{h}$.
In LoRA \citep{hu2021lora}, we keep $\W_0$ fixed, and introduce a perturbation to the weight matrix, $\Delta \W$,
\begin{align}  \label{eq:lora}
  \mathbf{h} &= \W_0 \a + \Delta \W \a = \W_0 \a + \B \A \a.
\end{align}
Critically, $\Delta \W$ is low-rank as it is written as the product of two matrices, $\B \in \mathbb{R}^{\nout\times\nlr}$ and $\A\in\mathbb{R}^{\nlr\times \nin}$ where $\nlr$ is significantly smaller than $\nin$ or $\nout$ (e.g. 4096), for instance, we use $\nlr=8$.
Therefore, the total number of LoRA parameters for this weight matrix is $\nlr (\nin + \nout)$, which is typically far smaller than the underlying number of parameters in the full matrix, $\nin \nout$. LoRA has shown great success in fine-tuning large scale models efficiently, and can be adapted to language or vision architectures \citep{peft}.
Note that one of the key motivations for introducing LoRA for LLM finetuning was the huge memory cost of maintaining the average gradient and average squared gradients of the e.g.\ 7 billion parameters in the optimizer.
This multiplies the memory required by a factor of 3, relative to the memory required to just load the weights.
LoRA massively reduces this memory cost to only 3 times the number of parameters in the LoRA adapters, as it only optimizes the LoRA adapters.

\subsection{Laplace approximations}
In Bayesian inference for classification or next token prediction, the goal is to find the full posterior,
\begin{align}
  \Pc{\param}{\X, \y} &\propto \Pc{\y}{\X, \param} \P{\param}.
\end{align}
Here, $\X\in\mathcal{T}^{N\times S}$ is the input, where $\mathcal{T}$ is the set of possible tokens, $N$ is the number of sequences, $S$ is the (max) sequence length.
The targets are denoted $\y\in\mathcal{Y}^{N}$, where $\mathcal{Y}$ could be different from $\mathcal{T}$ (e.g.\ in sentiment classification) or could be the same as $\mathcal{T}$ (e.g.\ for next token prediction).
Further, $\Pc{\param}{\X, \y}$ is the posterior, $\Pc{\y}{\param, \X}$ is the likelihood (e.g. softmax Categorical distribution for classification tasks). We use an isotropic Gaussian prior, with precision $\lambda$, 
\begin{align}
  \P{\param} &= \mathcal{N}(\bm{0}, \lambda^{-1} \I).
\end{align}
Calculating this posterior is usually intractable.
The Laplace approximation begins by finding the maximum a-posteriori (MAP) solution \citep{mackay1992practical} (i.e.\ the maximum of the log-joint, $\L{\param}$),
\begin{align}
  \L{\param} &= \log \Pc{\y}{\X, \param} + \log \P{\param} = \log \Pc{\param}{\X, \y} + \text{const}\\
  \paramMAP &= \argmax_{\param} \L{\param}.
\end{align}
Then the Laplace approximation consists of a second-order Taylor expansion of the log-joint around $\paramMAP$,
\begin{align} \label{eq:taylor}
  \L{\param} &\approx \L{\paramMAP} - \frac{1}{2} (\param-\paramMAP)^T (\nabla_{\param}^2 \L{\param}|_\paramMAP) (\param-\paramMAP).
\end{align}
Since the log-joint is now a quadratic function of $\param$, the approximate posterior becomes a Gaussian centered at $\param_\text{MAP}$ with covariance given by the inverse of the Hessian,
\begin{align} \label{eq:posterior}
    \Pc{\param}{\data} &\approx \N{\param; \param_\text{MAP}, \S},\\
    \S &= -(\nabla_\param^2 \L{\param}|_\paramMAP)^{-1} = -(\nabla_\param^2 \log \Pc{\y}{\X, \param}|_{\param_\text{MAP}} + \lambda \I)^{-1}.
\end{align}
To ensure positive definiteness of the covariance, the general approach is to transform gradients into estimates of the Hessian using either the Fisher information or the Generalized Gauss Newton (GGN) matrix \citep[see][for further details]{kunstner2019limitations}. We use the Fisher information, 
\begin{align}
    \F(\param) = \sum_{n=1}^N \mathbb{E}_{\Pc{y}{f_\param(\x_n)}} \bigg[ \nabla_\param \Pc{y}{f_\param(\x_n)}  \big(\nabla_\param \Pc{y}{f_\param(\x_n)} \big)^T \bigg],
\end{align}
where the expectation is taken over the model's output distribution.

Importantly, the full Hessian or Fisher is a $P \times P$ matrix, where $P$ is the number of parameters.
For LLMs, this might translate to a matrix with dimensions of 7 billion by 7 billion, which is clearly intractable. 
Even if we consider the Hessian only over the low-rank adapters this matrix is still too large; for example, applying rank 8 LoRA on Llama2-7B still yields roughly 6 million trainable parameters.
Consequently, we follow the usual Laplace approximation approach by imposing further structure on the Hessian.
In particular, we consider either just the Hessian for the last-layer, or using Kronecker-factored (KFAC) structure for individual weight matrices \citep{ritter2018scalable,daxberger2021laplace}.
In KFAC, we approximate the Fisher using blocks for each linear layer.
For the $\ell$th linear layer, we compute the block by denoting  the input as $\a_\ell$ and the output as $\mathbf{b}_\ell$. 
Then, the Fisher is,
\begin{align} \label{eq:kfac}
    \F_\ell = \sum_{n=1}^N \mathbb{E}_{\Pc{y}{f_\param(\x_n)}} \big[(\a_{\ell-1}\a_{\ell-1}^T)  \otimes (\g_{\ell}\g_{\ell}^T) \big].
\end{align} 
where $\g_\ell=\nabla_{\mathbf{b}_\ell} \log \Pc{\y}{\X, \param}$ is the gradient of the the log-likelihood gradient with respect to the outputs.

Using Laplace approximations has strong connections to linearizing the neural network \citep{kunstner2019limitations,immer2021improving}. 
As such, it is commonly found that predicting under the linearized model is more effective than e.g.\ sampling the approximate posterior over weights \citep{foong2019between,daxberger2021laplace,deng2022accelerated,antoran2022adapting}, (see Appendix~\ref{app:laplace_approx} for further details).
In particular,
\begin{align} \label{eq:linear}
    f_\param(\x_*) \approx f_{\param_\text{MAP}}(\x_*) + \nabla_\param f_{\param}(\x_*)|^T_{\param_\text{MAP}} (\param - \param_\text{MAP}).
\end{align}
where $\x_*$ is a test-input. This approach is also known as the linearized Laplace approximation.

Since we have the approximated posterior in Eq.~\eqref{eq:posterior} and the linearized model in Eq.~\eqref{eq:linear}, we can integrate out the posterior on weights and get a Gaussian posterior on output logits,
\begin{align} \label{eq:prediction}
    f_\param(\x_*) \sim \N{f_{\param_\text{MAP}}(\x_*),\mL},
\end{align}
where
\begin{align}
    \mL = (\nabla_\param f_{\param}(\x_*)|^T_{\param_\text{MAP}}) \S (\nabla_\param f_{\param}(\x_*)|_{\param_\text{MAP}}).
\end{align}
Subsequently, we can optimize the prior precision $\lambda$ using the closed form Laplace marginal likelihood (model evidence) \citep{immer2021improving,daxberger2021laplace} on the training dataset,
\begin{align}
  \label{eq:model_evidence}
    \Pc{\y}{\X} = \int \Pc{\y}{\X, \param} \P{\param} d\param \approx \exp(\L{\paramMAP}) (2\pi)^{D/2} |\S|^{1/2},
\end{align}
where we have applied the Taylor approximation in Eq.~\eqref{eq:taylor}. 
Crucially, unlike other post-hoc calibration methods, post-hoc Laplace does not require a separate validation set. 
This feature is particularly beneficial for small-scale datasets where training data is scarce.
To obtain samples of $f_\param(\x_*)$, we can decompose the covariance using the Cholesky factorization, $\mL =  \Lm\Lm^T$,
\begin{align}
  \label{eq:noisy_pred}
  \tilde{f}_\param(\x_*) = f_{\param_\text{MAP}}(\x_*) + \Lm \boldsymbol{\xi},
\end{align}
where $\boldsymbol{\xi}$ is a vector of IID standard normal random variables.  
We compute the Bayesian model average by computing the average probabilities (passing the sampled logits through softmax function) under the Gaussian random noise from $\boldsymbol{\xi}$.
There are common approaches to approximating the Bayesian model average arising from Eq.~\eqref{eq:noisy_pred}, including the generalized probit approximation and the Laplace bridge \citep{daxberger2021laplace,kristiadi2020being,Lu2020UncertaintyEW,mackay1998choice}, but we find that they perform considerably worse than naive Monte-Carlo sampling, potentially due to their ignorance of covariance terms (Appendix~\ref{app:laplace_approx}).

\section{Methods}

\newcommand{\w}{\mathbf{w}}
\newcommand{\bv}{\mathbf{b}}
\newcommand{\dw}{\mathbf{\Delta w}}
\newcommand{\wmap}{\w_\text{MAP}}
\newcommand{\const}{\text{const}}

\newcommand{\DlikeA}{\mathbf{D}_\text{like; a}}
\newcommand{\DlikeB}{\mathbf{D}_\text{like; b}}

\newcommand{\DpostA}{\mathbf{D}_\text{post; a}}
\newcommand{\DpostB}{\mathbf{D}_\text{post; b}}

\newcommand{\M}{\mathbf{M}}
\newcommand{\bL}{\mathbf{L}}
\newcommand{\K}{\mathbf{K}}
\newcommand{\D}{\mathbf{D}}
\newcommand{\U}{\mathbf{U}}
\newcommand{\V}{\mathbf{V}}
\newcommand{\na}{n_\text{a}}
\newcommand{\nb}{n_\text{b}}

\newcommand{\nlora}{n_\text{lora}}
\newcommand{\nkfac}{n_\text{kfac}}

\newcommand{\Sa}{\S_\text{a}}
\newcommand{\Sb}{n_\text{b}}

\newcommand{\Sprior}{\S_\text{prior}}
\newcommand{\Slike}{\S_\text{like}}
\newcommand{\Spost}{\S_\text{post}}

\newcommand{\SpriorA}{\S_\text{prior; a}}
\newcommand{\SlikeA}{\S_\text{like; a}}
\newcommand{\SpostA}{\S_\text{post; a}}

\newcommand{\SpriorB}{\S_\text{prior; b}}
\newcommand{\SlikeB}{\S_\text{like; b}}
\newcommand{\SpostB}{\S_\text{post; b}}

\newcommand{\0}{\mathbf{0}}

\renewcommand{\vec}{\operatorname{vec}}
%
%
%
%

Laplace-LoRA uses post-hoc Laplace Kronecker-factored approximations for the LoRA adapters. Specifically, we treat the adapter $\B \A \x$ in Eq.~\eqref{eq:lora} as two separate linear layers with weights $\A\in\mathbb{R}^{\nlr\times \nin}$ and $\B\in\mathbb{R}^{\nout\times \nlr}$ respectively rather than as a single linear layer with a low rank weight matrix. 
One issue with this approach is that the LoRA adapters are $\nlora \times d$ or $d \times \nlora$ dimensional, so while one Kronecker-factor is small ($\nlora \times \nlora$), the other is large ($d \times d$, where $d=4096$ in LlaMA2-7B attention layers).
These $d \times d$ Kronecker factors are roughly the same size as the underlying weight matrices, so representing them explicitly would eliminate any memory benefits of working with LoRA adapters.
As such, we are forced to use a low-rank representation of this Kronecker factor.
Specifically, we use a rank-$\nkfac$ representation (the rank of this low-rank approximation to the Kronecker factor will in general be different from the rank of the LoRA adapters).

To ensure that we retain memory efficiency at all steps, we need to be able to:
\begin{enumerate}
  \item Compute the low-rank approximation to the Kronecker factor ``incrementally'' (i.e.\ all the computations are low-rank and we do not e.g.\ compute the full-rank factor first, and then find a low-rank form; Appendix~\ref{app:lr_incremental}).
  \item Optimize the marginal likelihood using the low-rank approximation (Appendix~\ref{app:lr_marg_like}).
  \item Low-rank linearized prediction (Appendix~\ref{app:lr_linearised}).
\end{enumerate}

\section{Results}

We consider post-hoc Laplace approximations applied to LoRA parameters (Laplace-LoRA) at model checkpoints $\param_\text{MAP}$ obtained from standard fine-tuning.
In particular, we used the PEFT library \citep{peft} for fine-tuning LlaMA2-7B \citep{touvron2023llama2} on common-sense reasoning tasks (multiple choice or True/False classification). We applied LoRA to queries, values, and the output layer, using default hyperparameters from HuggingFace \citep{wolf2020transformers,peft} (see Appendix~\ref{app:hyperparam}). The fine-tuning was carried out with a batch size of 4 for 10,000 iterations. For True/False or multiple choice questions we selected the next token logits corresponding to True/False or A/B/C/D depending on each dataset (refer to Appendix~\ref{app:prompt} for our prompt templates), and fine-tuned the LLM to maximize the likelihood of the correct token. 


Model checkpoints were saved every 1000 gradient steps, and at each checkpoint, we applied post-hoc Laplace-LoRA with KFAC; we predicted using the linearized model as described in the Methods section.
For clarity, we refer to the full Laplace-LoRA approximation (which is applied to all LoRA fine-tuned weights) as LA, and the last-layer Laplace-LoRA approximation (targeting only the LoRA weights at the output layer to emulate standard last-layer Laplace approximation \citep{kristiadi2020being,daxberger2021laplace} ) as LLLA.

We assessed the efficacy of Laplace-LoRA by evaluating the negative log-likelihood and expected calibration error of LlaMA2-7B during fine-tuning on common-sense reasoning tasks. These metrics are further discussed in Appendix~\ref{app:metrics}.  We also drew comparisons with established baselines including Monte-Carlo dropout \citep{gal2016dropout} with an ensemble size of 10 (with a dropout rate of 0.1 during fine-tuning), checkpoint ensemble \citep{chen2017checkpoint} with an ensemble size of 3 using the most recent 3 checkpoints when available, deep ensemble \citep{lakshminarayanan2017simple,wang2023lora,zhai2023uncertainty} with three LoRA fine-tuned LLMs, and temperature scaling \citep{guo2017calibration}.
Since achieving good calibration and accurate uncertainty estimation is significantly more challenging on small-scale datasets \citep{zhao2020role}, our focus lies primarily on datasets comprising fewer than 10,000 training examples. 
We use the publicly available validation split across all benchmarks as the test set to evaluate performances at checkpoints.


\begin{figure*}[t]
    \centering
    \includegraphics[width=\textwidth]{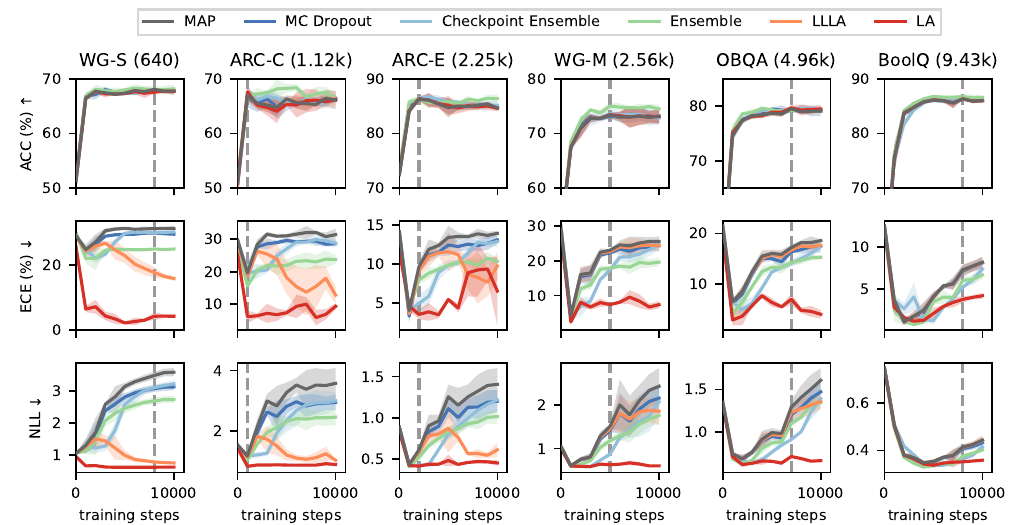}
    \caption{Fine-tuning of LlaMA2-7B across six common sense reasoning tasks (presented column-wise, with number of training examples in brackets), evaluated on the test set every 1000 gradient steps. The vertical dashed line gives the number of training steps with optimal MAP performance, and indicates that Laplace is likely to offer benefits even when combined with early stopping. MAP: standard fine-tuning; MC Dropout: Monte-Carlo dropout; Checkpoint Ensemble: ensembling three most recent checkpoints; Ensemble: ensembling three LoRA fine-tuned models; LLLA: last-layer Laplace approximation on LoRA weights in the output layer; LA: full Laplace approximation on all LoRA weights.}
    \label{fig:llama7b-kron}
\end{figure*}

\subsection{In-distribution fine-tuning and evaluation}
We began our evaluation with in-distribution fine-tuning on the following common sense reasoning tasks: 
Winogrande-small (WG-S), Winogrande-medium (WG-M) \citep{sakaguchi2021winogrande}, ARC-Challenge (ARC-C), ARC-Easy (ARC-E) \citep{allenai:arc}, openbook QA (OBQA) \citep{OpenBookQA2018}, and BoolQ \citep{clark2019boolq} benchmarks\footnote{We truncate the context in BoolQ for computational considerations (see Appendix~\ref{app:hyperparam}).}.
One of the key advantages of Laplace approximations is that they give an approximation of the model evidence which can be used to tune hyperparameters on the training set alone, without needing a separate validation set.
Thus, we first considered a setting with only a training set and a test set, with no distinct validation set for hyperparameter tuning. For our Laplace-LoRA methods (LA and LLLA), we optimized the prior variances using the Laplace model evidence (detailed in Algorithm~\ref{alg:LA_evidence} in Appendix~\ref{app:laplace_prior}).

Figure~\ref{fig:llama7b-kron} presents the accuracy (ACC), expected calibration error (ECE) and negative log-likelihood (NLL) on the evaluation set for each fine-tuned checkpoint. Since all other methods maintained similar ACC compared to MAP, our primary concern revolved around ECE and NLL. Notably, the very large NLLs for standard (MAP) fine-tuning as training proceeds indicates that overconfidence remains an important issue when when fine-tuning Llama-7B. Monte-Carlo dropout offers marginal improvements over MAP, while an checkpoint ensemble offers slightly more gains at the beginning. LoRA ensemble sometimes offer an improvement in ACC, but its improvements on ECE and NLL are always limited. Conversely, LLLA seems to give substantial improvements on some datasets (WG-S, ARC-C, ARC-E, but little if any improvement on others WG-M, OBQA, BoolQ). In contrast, LA consistently offers dramatic benefits in terms of ECE and NLL, indicating overconfidence on the evaluation set is dramatically reduced. 

Additionally, we present these results in tabular form in Table~\ref{table:id_train}.
We stopped at 5000 steps to somewhat mirror the effect of early stopping (note however that choosing the correct point to stop is difficult if not impossible to determine in this setting which lacks a validation set.)   LA consistently offered considerable gains over MAP in terms of ECE and NLL, achieving best NLL among all methods across all datasets, while maintaining similar ACC.
Note that while this table gives only the performance at one time point, Figure~\ref{fig:llama7b-kron} shows that LA usually offers considerable benefits in calibration across all points in training. Additionally, in Fig.~\ref{fig:llama7b-kron}, we highlight the point with highest \textit{test} accuracy with the dashed line (as this is the highest test accuracy, it cannot practically be used to early-stop, but it does indicate that were we to optimally early-stop, we would still get dramatic benefits in calibration from LA.)

\begin{table}[t]
\small
\centering
\caption{Comparison of different post-hoc methods applied to the fine-tuned LlaMA2-7B across six common sense reasoning tasks. Results are evaluated at the early stopping point of 5000 gradient steps. We report standard deviations in subscripts, and bold numbers that are statistical significant.}
\label{table:id_train}
\vspace{0.1cm}
\begin{tabular}{l|| l |l l l l l l}
Metrics & Methods & WG-S & ARC-C & ARC-E & WG-M & OBQA & BoolQ \\
\hline
\hline
\multirow{5}{*}{ACC $\uparrow$} \rule{0pt}{2.25ex}
& MAP    & $67.4_{0.3}$ & $66.3_{0.6}$ & $84.7_{1.5}$ & $73.4_{0.4}$ & $78.7_{0.4}$ & $86.1_{0.2}$ \\
& MC Drop & $67.8_{0.1}$ & $65.3_{1.0}$ & $85.0_{1.3}$ & $73.2_{0.5}$ & $79.5_{0.2}$ & $86.0_{0.3}$ \\
& Ckpt Ens   & $67.4_{0.2}$ & $65.5_{0.4}$ & $85.8_{0.2}$ & $73.6_{0.7}$ & $79.1_{0.1}$ & $86.3_{0.2}$ \\
& Ensemble        & $68.0_{0.3}$ & $68.2_{0.7}$ & $85.8_{0.5}$ & $75.0_{0.5}$ & $79.3_{0.4}$ & $86.8_{0.1}$ \\
\cdashline{2-8} \rule{0pt}{2.25ex}
& LLLA   & $67.4_{0.3}$ & $66.2_{0.4}$ & $84.7_{1.5}$ & $73.4_{0.4}$ & $78.7_{0.4}$ & $86.1_{0.2}$ \\
& LA     & $67.3_{0.2}$ & $65.3_{0.2}$ & $85.1_{1.5}$ & $73.4_{0.3}$ & $78.9_{0.2}$ & $86.1_{0.2}$ \\
\hline
\multirow{5}{*}{ECE $\downarrow$} \rule{0pt}{2.25ex}
& MAP    & $31.2_{0.3}$ & $31.0_{0.5}$ & $13.4_{1.3}$ & $23.0_{0.1}$ & $16.1_{0.6}$ & $4.0_{0.5}$ \\
& MC Drop    & $29.4_{0.3}$ & $29.6_{0.8}$ & $12.4_{1.2}$ & $22.2_{0.2}$ & $15.0_{0.4}$ & $4.1_{0.4}$ \\
& Ckpt Ens   & $29.7_{0.6}$ & $27.0_{1.5}$ & $9.8_{0.6}$ & $17.4_{0.9}$ & $12.4_{0.3}$ & $\bm{1.2_{0.4}}$ \\
& Ensemble        & $24.7_{0.3}$ & $21.9_{1.7}$ & $9.9_{0.2}$ & $17.9_{0.6}$ & $13.3_{0.6}$ & $3.5_{0.2}$ \\
\cdashline{2-8} \rule{0pt}{2.25ex}
& LLLA   & $22.8_{2.0}$ & $18.2_{4.4}$ & $11.6_{2.2}$ & $22.6_{0.2}$ & $15.8_{0.6}$ & $4.0_{0.5}$ \\
& LA     & $\bm{2.1_{0.3}}$ & $\bm{7.4_{0.7}}$ & $\bm{5.4_{0.2}}$ & $\bm{7.4_{0.4}}$ & $\bm{6.4_{0.8}}$ & $2.1_{0.6}$ \\
\hline
\multirow{5}{*}{NLL $\downarrow$} \rule{0pt}{2.25ex}
& MAP    & $3.15_{0.10}$ & $3.28_{0.29}$ & $1.26_{0.13}$ & $1.51_{0.05}$ & $0.99_{0.05}$ & $0.35_{0.01}$ \\
& MC Drop     & $2.81_{0.11}$ & $2.82_{0.21}$ & $1.11_{0.10}$ & $1.41_{0.03}$ & $0.95_{0.04}$ & $0.35_{0.01}$ \\
& Ckpt Ens   & $2.58_{0.15}$ & $2.36_{0.34}$ & $0.80_{0.06}$ & $0.87_{0.06}$ & $0.76_{0.01}$ & $\bm{0.33_{0.00}}$ \\
& Ensemble        & $2.46_{0.14}$ & $2.32_{0.14}$ & $0.83_{0.06}$ & $1.18_{0.05}$ & $0.87_{0.03}$ & $0.34_{0.00}$ \\
\cdashline{2-8} \rule{0pt}{2.25ex}
& LLLA   & $0.98_{0.13}$ & $1.21_{0.16}$ & $0.87_{0.26}$ & $1.45_{0.06}$ & $0.97_{0.04}$ & $0.35_{0.01}$ \\
& LA     & $\bm{0.60_{0.01}}$ & $\bm{0.88_{0.03}}$ & $\bm{0.49_{0.06}}$ & $\bm{0.63_{0.02}}$ & $\bm{0.65_{0.01}}$ & $0.34_{0.01}$ \\
\end{tabular}
\end{table}

Next, we considered a more standard setting, in which the training set was partitioned into an 80\% subset for training and a 20\% validation subset. 
Notably, this usually led to a slight decrease in performance, as we were no longer training on all the data (Table~\ref{table:id_train} against Table~\ref{table:id_val}).
This allowed us to benchmark against a widely-adopted post-hoc calibration technique, temperature scaling \citep{guo2017calibration}, which requires a separate validation set to tune the temperature hyperparameter. 
Additionally, it is possible to tune the prior precision in Laplace using the NLL on the validation set, rather than the ``more Bayesian'' approach of tuning prior precision using the Laplace estimate of the model evidence. 
The standard approach is to perform grid search to maximize validation log-likelihood \citep{daxberger2021laplace}. 
However, fine-grained grid search is costly. Instead, utilizing our reparametrized model average (Eq.~\ref{eq:noisy_pred}), weperformed stochastic gradient descent for the precision hyperparameters, using the validation NLL as the objective (see Algorithm~\ref{alg:LA_val} in Appendix~\ref{app:laplace_prior}).
Table ~\ref{table:id_val} shows the result of tuning temperature and Laplace prior on a validation set at the checkpoint with the best MAP validation accuracy. Although temperature scaling is very competitive, LA still gives slightly improved NLL and usually improves ECE.

\begin{table}[t]
\small
\centering
\caption{Comparison of different post-hoc methods applied to the fine-tuned LlaMA2-7B across six common sense reasoning tasks, with a validation set split from the training set used for tuning temperature and Laplace prior precision. Results are evaluated at the best MAP performance checkpoint observed on the validation set.}
\label{table:id_val}
\vspace{0.1cm}
\begin{tabular}{l|| l |l l l l l l}
 Metrics & Methods & WG-S & ARC-C & ARC-E & WG-M & OBQA & BoolQ \\
\hline \hline
\multirow{6}{*}{ACC $\uparrow$} \rule{0pt}{2.25ex}
& MAP    & $67.0_{0.6}$ & $64.9_{1.1}$ & $85.2_{0.6}$ & $73.7_{0.9}$ & $77.7_{0.8}$ & $85.8_{0.4}$ \\
& MC Drop     & $66.7_{0.3}$ & $64.9_{1.9}$ & $85.1_{0.5}$ & $73.5_{0.9}$ & $77.7_{0.2}$ & $85.9_{0.4}$ \\
& Ckpt Ens     & $66.7_{0.3}$ & $64.9_{1.1}$ & $85.2_{0.6}$ & $73.8_{1.0}$ & $78.2_{0.2}$ & $85.4_{0.3}$ \\
& Temp   & $67.0_{0.6}$ & $64.9_{1.1}$ & $85.2_{0.6}$ & $73.7_{0.9}$ & $77.7_{0.8}$ & $85.8_{0.4}$ \\
\cdashline{2-8} \rule{0pt}{2.25ex}
& LLLA   & $66.9_{0.5}$ & $66.1_{0.6}$ & $84.8_{0.5}$ & $73.7_{0.9}$ & $77.6_{0.7}$ & $85.8_{0.4}$ \\
& LA     & $66.9_{0.6}$ & $66.9_{1.1}$ & $85.4_{0.4}$ & $73.7_{1.0}$ & $78.1_{0.7}$ & $85.8_{0.4}$ \\
\hline
\multirow{6}{*}{ECE $\downarrow$} \rule{0pt}{2.25ex}
& MAP    & $30.8_{1.8}$ & $26.1_{1.4}$ & $8.9_{0.3}$ & $24.9_{1.3}$ & $9.8_{1.0}$ & $7.4_{0.1}$ \\
& MC Drop     & $29.5_{1.6}$ & $25.6_{0.7}$ & $8.8_{0.6}$ & $23.5_{1.2}$ & $8.8_{0.8}$ & $7.5_{0.1}$ \\
& Ckpt Ens     & $25.2_{1.6}$ & $26.1_{1.4}$ & $8.9_{0.3}$ & $22.8_{1.4}$ & $4.7_{0.5}$ & $3.2_{0.5}$ \\
& Temp   & $12.8_{0.9}$ & $\bm{4.6_{1.0}}$ & $4.7_{0.8}$ & $6.3_{1.6}$ & $7.2_{2.6}$ & $2.5_{0.3}$ \\
\cdashline{2-8} \rule{0pt}{2.25ex}
& LLLA   & $11.6_{1.3}$ & $5.6_{2.1}$ & $4.2_{0.3}$ & $\bm{3.8_{1.4}}$ & $5.4_{0.4}$ & $\bm{1.7_{0.5}}$ \\
& LA     & $\bm{7.8_{1.9}}$ & $7.5_{1.2}$ & $\bm{3.4_{0.8}}$ & $4.8_{1.6}$ & $\bm{3.5_{0.4}}$ & $1.9_{0.3}$ \\
\hline
\multirow{6}{*}{NLL $\downarrow$} \rule{0pt}{2.25ex}
& MAP    & $2.75_{0.57}$ & $1.64_{0.19}$ & $0.54_{0.03}$ & $2.43_{0.50}$ & $0.71_{0.03}$ & $0.43_{0.01}$ \\
& MC Drop     & $2.54_{0.49}$ & $1.55_{0.16}$ & $0.52_{0.04}$ & $2.12_{0.35}$ & $0.71_{0.04}$ & $0.43_{0.01}$ \\
& Ckpt Ens     & $1.31_{0.04}$ & $1.64_{0.18}$ & $0.54_{0.03}$ & $1.89_{0.24}$ & $0.65_{0.02}$ & $0.35_{0.01}$ \\
& Temp   & $0.68_{0.01}$ & $0.90_{0.01}$ & $0.43_{0.02}$ & $0.58_{0.01}$ & $0.67_{0.02}$ & $0.35_{0.00}$ \\
\cdashline{2-8} \rule{0pt}{2.25ex}
& LLLA   & $0.68_{0.01}$ & $0.94_{0.02}$ & $0.44_{0.01}$ & $0.56_{0.01}$ & $0.66_{0.02}$ & $0.35_{0.00}$ \\
& LA     & $\bm{0.66_{0.02}}$ & $\bm{0.86_{0.02}}$ & $\bm{0.41_{0.02}}$ & $\bm{0.55_{0.01}}$ & $\bm{0.62_{0.01}}$ & $\bm{0.34_{0.00}}$ \\
\end{tabular}
\end{table}

\subsection{Evaluations under distribution shift}
Finally, since real world deployment of LLMs requires them to generalize to data coming from a different domain compared to the fine-tuning dataset \citep{ouyang2022training,touvron2023llama,touvron2023llama2}, we conduct further experiments to evaluate fine-tuned models under distribution shift using the model checkpoint fine-tuned on OBQA in Table~\ref{table:id_val}. Our evaluations under distribution shift are divided into two groups, smaller distribution shift and larger distribution shift. For smaller distribution shift, we use ARC datasets ARC-C and ARC-E, as they also contain common sense reasoning multiple choice questions similar to those in OBQA. For larger distribution shift, we pick four subjects from the MMLU task \citep{hendrycks2020measuring}: computer science (CS), engineering (Eng), law, and health, where each subject contains one or more tasks. A detailed breakdown is available in Table~\ref{table:mmlu} in Appendix~\ref{app:mmlu}.

Table~\ref{table:ood_train} shows the OOD evaluation results using the early-stopped checkpoint on OBQA from Table~\ref{table:id_train}. LA clearly provides substantial improvements over all other methods in terms of Expected Calibration Error (ECE) and Negative Log-Likelihood (NLL) while maintaining similar accuracy (ACC), with a slight improvement in the Computer Science (CS) subject.
Table~\ref{table:ood_val} shows the evaluation results under distribution shift using the best validation accuracy checkpoint on OBQA from Table~\ref{table:id_val}. 
Once again, both LLLA and LA typically yield improvements in ECE and NLL over other methods, indicating their effectiveness in out-of-distribution scenarios.

\begin{table}[t] 
\small
\centering
\caption{Comparison of different post-hoc methods across six out-of-distribution datasets. Results are evaluated using the fine-tuned LlaMA2-7B on OBQA as in Table~\ref{table:id_train}.}
\label{table:ood_train}
\vspace{0.1cm}
\resizebox{0.975\textwidth}{!}{\begin{tabular}{l|| l | l |l l | l l l l}

&  & \multicolumn{1}{c|}{ID} & \multicolumn{2}{c|}{Smaller Distribution Shift} & \multicolumn{4}{c}{Larger Distribution Shift} \\
\rule{0pt}{2.25ex}
Metrics & Methods & OBQA & ARC-C & ARC-E & CS & Eng & Law & Health \\
\hline\hline
\multirow{6}{*}{ACC $\uparrow$} \rule{0pt}{2.25ex}
& MAP    & $78.7_{0.4}$ & $67.9_{1.4}$ & $77.7_{0.3}$ & $42.0_{3.2}$ & $41.2_{2.0}$ & $37.4_{0.4}$ & $48.3_{0.3}$ \\
& MC Drop  & $79.5_{0.2}$ & $67.7_{0.6}$ & $77.2_{0.6}$ & $41.9_{2.2}$ & $39.6_{1.7}$ & $37.9_{0.4}$ & $48.2_{0.9}$ \\
& Ckpt Ens     & $79.1_{0.2}$ & $67.9_{0.8}$ & $77.4_{0.8}$ & $41.1_{2.0}$ & $38.7_{1.2}$ & $37.7_{0.2}$ & $48.2_{0.6}$ \\
& Ensemble   & $68.0_{0.3}$ & $69.0_{0.7}$ & $77.0_{1.1}$ & $41.9_{1.7}$ & $41.9_{2.8}$ & $38.0_{0.3}$ & $48.4_{0.5}$ \\
\cdashline{2-9} \rule{0pt}{2.25ex}
& LLLA   & $78.7_{0.4}$ & $68.1_{0.0}$ & $78.1_{0.0}$ & $45.6_{0.0}$ & $38.9_{0.0}$ & $37.1_{0.0}$ & $48.5_{0.0}$ \\
& LA     & $78.9_{0.2}$ & $69.2_{0.0}$ & $78.5_{0.0}$ & $45.1_{0.0}$ & $39.1_{0.0}$ & $37.3_{0.0}$ & $49.1_{0.0}$ \\
\hline
\multirow{6}{*}{ECE $\downarrow$} \rule{0pt}{2.25ex}
& MAP    & $16.1_{0.6}$ & $22.2_{1.2}$ & $15.8_{1.0}$ & $34.2_{3.1}$ & $38.4_{1.7}$ & $35.2_{0.7}$ & $34.2_{0.8}$ \\
& MC Drop  & $15.0_{0.4}$ & $21.4_{0.4}$ & $15.5_{1.0}$ & $33.7_{2.0}$ & $38.6_{2.9}$ & $34.2_{0.6}$ & $33.5_{0.2}$ \\
& Ckpt Ens     & $10.1_{0.3}$ & $17.7_{0.7}$ & $12.1_{0.6}$ & $29.1_{2.3}$ & $32.5_{1.8}$ & $32.1_{0.1}$ & $29.0_{0.3}$ \\
& Ensemble   & $24.7_{0.3}$ & $18.5_{1.0}$ & $13.4_{0.7}$ & $30.4_{2.6}$ & $31.7_{1.3}$ & $32.7_{0.4}$ & $31.0_{0.9}$ \\
\cdashline{2-9} \rule{0pt}{2.25ex}
& LLLA   & $15.8_{0.6}$ & $21.3_{0.0}$ & $14.8_{0.0}$ & $30.3_{0.0}$ & $39.7_{0.0}$ & $33.6_{0.0}$ & $33.5_{0.0}$ \\
& LA     & $\bm{6.4_{0.8}}$ & $\bm{8.8_{0.0}}$ & $\bm{6.2_{0.0}}$ & $\bm{15.8_{0.0}}$ & $\bm{25.5_{0.0}}$ & $\bm{24.7_{0.0}}$ & $\bm{17.9_{0.0}}$ \\
\hline
\multirow{6}{*}{NLL $\downarrow$} \rule{0pt}{2.25ex}
& MAP    & $0.99_{0.05}$ & $1.30_{0.07}$ & $1.04_{0.10}$ & $1.90_{0.12}$ & $2.19_{0.15}$ & $2.12_{0.03}$ & $2.09_{0.08}$ \\
& MC Drop  & $0.95_{0.04}$ & $1.24_{0.06}$ & $1.01_{0.09}$ & $1.86_{0.10}$ & $2.14_{0.13}$ & $2.09_{0.02}$ & $2.05_{0.07}$ \\
& Ckpt Ens     & $0.68_{0.03}$ & $1.03_{0.03}$ & $0.80_{0.03}$ & $1.55_{0.04}$ & $1.72_{0.01}$ & $1.94_{0.01}$ & $1.74_{0.02}$ \\
& Ensemble   & $2.46_{0.14}$ & $1.15_{0.07}$ & $0.91_{0.04}$ & $1.74_{0.10}$ & $1.95_{0.10}$ & $2.04_{0.03}$ & $1.95_{0.11}$ \\
\cdashline{2-9} \rule{0pt}{2.25ex}
& LLLA   & $0.97_{0.04}$ & $1.30_{0.00}$ & $0.99_{0.00}$ & $1.80_{0.00}$ & $2.18_{0.00}$ & $2.06_{0.00}$ & $2.05_{0.00}$ \\
& LA     & $\bm{0.65_{0.01}}$ & $\bm{0.90_{0.00}}$ & $\bm{0.70_{0.00}}$ & $\bm{1.35_{0.00}}$ & $\bm{1.58_{0.00}}$ & $\bm{1.74_{0.00}}$ & $\bm{1.50_{0.00}}$ \\
\end{tabular}}
\end{table}

In addition to experiments on LlaMA in the main text, we also present results for fine-tuning encoder-based language models RoBERTa-base and RoBERTa-large \citep{liu2019roberta} on text classification tasks in Appendix~\ref{app:roberta}
Notably, Monte-Carlo dropout performs much better on RoBERTa-base but is less effective on RoBERTa-large; checkpoint ensemble often performs much worse than MAP in contrast to its performance on LlaMA; LLLA still offers tiny improvements over MAP. 
Besides, we have also added additional experiments on fine-tuning Mistral-7B \citep{jiang2023mistral} in Appendix~\ref{app:backbones}. Overall, full Laplace-LoRA (LA) provides consistent and significant improvements in uncertainty estimation across models (RoBERTa, LlaMA, Mistral) and tasks (text classification, common sense reasoning).
Supplementary experiments using diagonal Fisher as a Hessian approximation instead of KFAC are documented in Appendix~\ref{app:diag}. However, we found that diagonal Laplace approximations do not offer consistent improvements in ECE and NLL, emphasizing the importance of using KFAC to model correlations between weights in Laplace-LoRA. 
We have open sourced our code implementations (Appendix~\ref{app:code}).


\subsection{Memory and runtime cost}

The primary advantages of LoRA lie in its memory and computational efficiency. In this section, we present a comparative analysis of the memory and time costs associated with LoRA fine-tuning alone, and LoRA fine-tuning followed by post-hoc application of Laplace-LoRA. As illustrated in Table~\ref{table:compute_cost}, implementing Laplace-LoRA incurs only an additional around 1--5\% memory overhead.  Note that in the experiments in Table~\ref{table:compute_cost}, there is very little overhead in terms of time (about 10\%), despite the fact that we actually accumulate the low-rank Kronecker factors 10 times (once at the end of every 1000 iterations, and we ran these experiments for 10000 iterations total).  In practice, we would only need to accumulate low-rank Kronecker factors once, so the overhead should be around 1\%.
This empirical analysis demonstrates that our low-rank KFAC approximation of Laplace-LoRA effectively maintains LoRA's efficiency.

\begin{table}
\small
\caption{Comparison of running time and memory cost of standard LoRA finetuning for 10,000 iterations vs standard LoRA finetuning for 10,000 along with accumulating low-rank Kronecker factors for Laplace every 1000 epochs. Experiements carried out using Llama2 7B, LoRA denotes standard LoRA fine-tuning for 10,000 steps, Laplace-LoRA denotes LoRA fine-tuning for 10,000 steps along with post-hoc Laplace approximation on all LoRA weights.
\label{table:compute_cost}}
\begin{tabular}{l||l|llllll}
   Metric & Method & WG-S & WG-M & ARC-C & ARC-E & OBQA & BoolQ \\
   \hline\hline
   \multirow{2}{*}{Time (Seconds) $\downarrow$} \rule{0pt}{2.25ex} & LoRA & 8287 & 8684 & 10500 &10301 & 9268 & 16964 \\
   & LoRA + Laplace & 8442 & 8839 & 11141 & 10942 & 10526 & 17636 \\
   \hline
   \multirow{2}{*}{Memory (GB) $\downarrow$} \rule{0pt}{2.25ex} & LoRA & 8.34 & 8.33 & 8.37 & 8.43 & 8.59 & 10.48 \\
   & LoRA + Laplace& 8.43 & 8.43 & 8.73 & 8.75 & 8.84 & 10.93 \\
\end{tabular}
\end{table}

\section{Conclusion}
In this work, we proposed Laplace-LoRA, for Bayesian parameter-efficient fine-tuning of LLMs. 
Our method is scalable to larger networks due to its focus on the LoRA weights. 
Additionally, by applying Laplace approximations post-hoc, after fine-tuning, we require no changes to efficient implementations of the standard fine-tuning process.
In our experiments, we observed significant gains in expected calibration error and negative log-likelihood, indicating an improved estimation of uncertainty. 
Our approach underpins the potential of Laplace-LoRA as a step towards the development of more reliable and trustworthy LLMs. 


\section{Acknowledgements}
This work was carried out using the computational facilities of the Advanced Computing Research Centre, University of Bristol - \url{http://www.bris.ac.uk/acrc/}.
We would like to thank Dr Stewart for funding compute resources used in this project, and we would like to thank Yixuan Su for helpful discussions.

\newpage
\bibliography{ref}

\begin{thebibliography}{72}
\providecommand{\natexlab}[1]{#1}
\providecommand{\url}[1]{\texttt{#1}}
\expandafter\ifx\csname urlstyle\endcsname\relax
  \providecommand{\doi}[1]{doi: #1}\else
  \providecommand{\doi}{doi: \begingroup \urlstyle{rm}\Url}\fi

\bibitem[Aitchison et~al.(2021)Aitchison, Yang, and Ober]{aitchison2021deep}
Laurence Aitchison, Adam Yang, and Sebastian~W Ober.
\newblock Deep kernel processes.
\newblock In \emph{International Conference on Machine Learning}, pp.\
  130--140. PMLR, 2021.

\bibitem[Antor{\'a}n et~al.(2022)Antor{\'a}n, Janz, Allingham, Daxberger,
  Barbano, Nalisnick, and Hern{\'a}ndez-Lobato]{antoran2022adapting}
Javier Antor{\'a}n, David Janz, James~U Allingham, Erik Daxberger, Riccardo~Rb
  Barbano, Eric Nalisnick, and Jos{\'e}~Miguel Hern{\'a}ndez-Lobato.
\newblock Adapting the linearised laplace model evidence for modern deep
  learning.
\newblock In \emph{ICML}, 2022.

\bibitem[Ashukha et~al.(2020)Ashukha, Lyzhov, Molchanov, and
  Vetrov]{ashukha2020pitfalls}
Arsenii Ashukha, Alexander Lyzhov, Dmitry Molchanov, and Dmitry Vetrov.
\newblock Pitfalls of in-domain uncertainty estimation and ensembling in deep
  learning.
\newblock \emph{arXiv preprint arXiv:2002.06470}, 2020.

\bibitem[Blundell et~al.(2015)Blundell, Cornebise, Kavukcuoglu, and
  Wierstra]{blundell2015weight}
Charles Blundell, Julien Cornebise, Koray Kavukcuoglu, and Daan Wierstra.
\newblock Weight uncertainty in neural network.
\newblock In \emph{International conference on machine learning}, pp.\
  1613--1622. PMLR, 2015.

\bibitem[Chen et~al.(2017)Chen, Lundberg, and Lee]{chen2017checkpoint}
Hugh Chen, Scott Lundberg, and Su-In Lee.
\newblock Checkpoint ensembles: Ensemble methods from a single training
  process.
\newblock \emph{arXiv preprint arXiv:1710.03282}, 2017.

\bibitem[Chen \& Li(2023)Chen and Li]{chen2023calibrating}
Wenlong Chen and Yingzhen Li.
\newblock Calibrating transformers via sparse gaussian processes.
\newblock \emph{arXiv preprint arXiv:2303.02444}, 2023.

\bibitem[Chung et~al.(2022)Chung, Hou, Longpre, Zoph, Tay, Fedus, Li, Wang,
  Dehghani, Brahma, et~al.]{chung2022scaling}
Hyung~Won Chung, Le~Hou, Shayne Longpre, Barret Zoph, Yi~Tay, William Fedus,
  Eric Li, Xuezhi Wang, Mostafa Dehghani, Siddhartha Brahma, et~al.
\newblock Scaling instruction-finetuned language models.
\newblock \emph{arXiv preprint arXiv:2210.11416}, 2022.

\bibitem[Cinquin et~al.(2021)Cinquin, Immer, Horn, and
  Fortuin]{cinquin2021pathologies}
Tristan Cinquin, Alexander Immer, Max Horn, and Vincent Fortuin.
\newblock Pathologies in priors and inference for bayesian transformers.
\newblock \emph{arXiv preprint arXiv:2110.04020}, 2021.

\bibitem[Clark et~al.(2019)Clark, Lee, Chang, Kwiatkowski, Collins, and
  Toutanova]{clark2019boolq}
Christopher Clark, Kenton Lee, Ming-Wei Chang, Tom Kwiatkowski, Michael
  Collins, and Kristina Toutanova.
\newblock Boolq: Exploring the surprising difficulty of natural yes/no
  questions.
\newblock In \emph{NAACL}, 2019.

\bibitem[Clark et~al.(2018)Clark, Cowhey, Etzioni, Khot, Sabharwal, Schoenick,
  and Tafjord]{allenai:arc}
Peter Clark, Isaac Cowhey, Oren Etzioni, Tushar Khot, Ashish Sabharwal, Carissa
  Schoenick, and Oyvind Tafjord.
\newblock Think you have solved question answering? try arc, the ai2 reasoning
  challenge.
\newblock \emph{arXiv:1803.05457v1}, 2018.

\bibitem[Daxberger et~al.(2021{\natexlab{a}})Daxberger, Kristiadi, Immer,
  Eschenhagen, Bauer, and Hennig]{daxberger2021laplace}
Erik Daxberger, Agustinus Kristiadi, Alexander Immer, Runa Eschenhagen,
  Matthias Bauer, and Philipp Hennig.
\newblock Laplace redux-effortless bayesian deep learning.
\newblock \emph{NeurIPS}, 2021{\natexlab{a}}.

\bibitem[Daxberger et~al.(2021{\natexlab{b}})Daxberger, Nalisnick, Allingham,
  Antor{\'a}n, and Hern{\'a}ndez-Lobato]{daxberger2021bayesian}
Erik Daxberger, Eric Nalisnick, James~U Allingham, Javier Antor{\'a}n, and
  Jos{\'e}~Miguel Hern{\'a}ndez-Lobato.
\newblock Bayesian deep learning via subnetwork inference.
\newblock In \emph{ICML}, 2021{\natexlab{b}}.

\bibitem[Deng et~al.(2022)Deng, Zhou, and Zhu]{deng2022accelerated}
Zhijie Deng, Feng Zhou, and Jun Zhu.
\newblock Accelerated linearized laplace approximation for bayesian deep
  learning.
\newblock \emph{NeurIPS}, 2022.

\bibitem[Dettmers et~al.(2023)Dettmers, Pagnoni, Holtzman, and
  Zettlemoyer]{dettmers2023qlora}
Tim Dettmers, Artidoro Pagnoni, Ari Holtzman, and Luke Zettlemoyer.
\newblock Qlora: Efficient finetuning of quantized llms.
\newblock \emph{arXiv preprint arXiv:2305.14314}, 2023.

\bibitem[Ding et~al.(2022)Ding, Qin, Yang, Wei, Yang, Su, Hu, Chen, Chan, Chen,
  et~al.]{ding2022delta}
Ning Ding, Yujia Qin, Guang Yang, Fuchao Wei, Zonghan Yang, Yusheng Su,
  Shengding Hu, Yulin Chen, Chi-Min Chan, Weize Chen, et~al.
\newblock Delta tuning: A comprehensive study of parameter efficient methods
  for pre-trained language models.
\newblock \emph{arXiv preprint arXiv:2203.06904}, 2022.

\bibitem[Ding et~al.(2023)Ding, Qin, Yang, Wei, Yang, Su, Hu, Chen, Chan, Chen,
  Yi, Zhao, Wang, Liu, Zheng, Chen, Liu, Tang, Li, and Sun]{opendelta}
Ning Ding, Yujia Qin, Guang Yang, Fuchao Wei, Zonghan Yang, Yusheng Su,
  Shengding Hu, Yulin Chen, Chi-Min Chan, Weize Chen, Jing Yi, Weilin Zhao,
  Xiaozhi Wang, Zhiyuan Liu, Hai-Tao Zheng, Jianfei Chen, Yang Liu, Jie Tang,
  Juanzi Li, and Maosong Sun.
\newblock Parameter-efficient fine-tuning of large-scale pre-trained language
  models.
\newblock \emph{Nature Machine Intelligence}, 5\penalty0 (3):\penalty0
  220--235, 2023.

\bibitem[Dusenberry et~al.(2020)Dusenberry, Jerfel, Wen, Ma, Snoek, Heller,
  Lakshminarayanan, and Tran]{dusenberry2020efficient}
Michael Dusenberry, Ghassen Jerfel, Yeming Wen, Yian Ma, Jasper Snoek,
  Katherine Heller, Balaji Lakshminarayanan, and Dustin Tran.
\newblock Efficient and scalable bayesian neural nets with rank-1 factors.
\newblock In \emph{International conference on machine learning}, pp.\
  2782--2792. PMLR, 2020.

\bibitem[Fan et~al.(2020)Fan, Zhang, Chen, and Zhou]{fan2020bayesian}
Xinjie Fan, Shujian Zhang, Bo~Chen, and Mingyuan Zhou.
\newblock Bayesian attention modules.
\newblock \emph{Advances in Neural Information Processing Systems},
  33:\penalty0 16362--16376, 2020.

\bibitem[Foong et~al.(2019)Foong, Li, Hern{\'a}ndez-Lobato, and
  Turner]{foong2019between}
Andrew~YK Foong, Yingzhen Li, Jos{\'e}~Miguel Hern{\'a}ndez-Lobato, and
  Richard~E Turner.
\newblock 'in-between'uncertainty in bayesian neural networks.
\newblock In \emph{ICML Workshop on Uncertainty and Robustness in Deep
  Learning}, 2019.

\bibitem[Fortuin et~al.(2021)Fortuin, Garriga-Alonso, Ober, Wenzel, R{\"a}tsch,
  Turner, van~der Wilk, and Aitchison]{fortuin2021bayesian}
Vincent Fortuin, Adri{\`a} Garriga-Alonso, Sebastian~W Ober, Florian Wenzel,
  Gunnar R{\"a}tsch, Richard~E Turner, Mark van~der Wilk, and Laurence
  Aitchison.
\newblock Bayesian neural network priors revisited.
\newblock \emph{arXiv preprint arXiv:2102.06571}, 2021.

\bibitem[Gal \& Ghahramani(2016)Gal and Ghahramani]{gal2016dropout}
Yarin Gal and Zoubin Ghahramani.
\newblock Dropout as a bayesian approximation: Representing model uncertainty
  in deep learning.
\newblock In \emph{international conference on machine learning}, pp.\
  1050--1059. PMLR, 2016.

\bibitem[Guo et~al.(2017)Guo, Pleiss, Sun, and Weinberger]{guo2017calibration}
Chuan Guo, Geoff Pleiss, Yu~Sun, and Kilian~Q Weinberger.
\newblock On calibration of modern neural networks.
\newblock In \emph{International conference on machine learning}, pp.\
  1321--1330. PMLR, 2017.

\bibitem[He et~al.(2023)He, Chen, and Zhu]{he2023preserving}
Guande He, Jianfei Chen, and Jun Zhu.
\newblock Preserving pre-trained features helps calibrate fine-tuned language
  models.
\newblock In \emph{ICLR}, 2023.

\bibitem[Hendrycks et~al.(2020)Hendrycks, Burns, Basart, Zou, Mazeika, Song,
  and Steinhardt]{hendrycks2020measuring}
Dan Hendrycks, Collin Burns, Steven Basart, Andy Zou, Mantas Mazeika, Dawn
  Song, and Jacob Steinhardt.
\newblock Measuring massive multitask language understanding.
\newblock \emph{arXiv preprint arXiv:2009.03300}, 2020.

\bibitem[Houlsby et~al.(2019)Houlsby, Giurgiu, Jastrzebski, Morrone,
  De~Laroussilhe, Gesmundo, Attariyan, and Gelly]{houlsby2019parameter}
Neil Houlsby, Andrei Giurgiu, Stanislaw Jastrzebski, Bruna Morrone, Quentin
  De~Laroussilhe, Andrea Gesmundo, Mona Attariyan, and Sylvain Gelly.
\newblock Parameter-efficient transfer learning for nlp.
\newblock In \emph{International Conference on Machine Learning}, pp.\
  2790--2799. PMLR, 2019.

\bibitem[Hu et~al.(2021)Hu, Shen, Wallis, Allen-Zhu, Li, Wang, Wang, and
  Chen]{hu2021lora}
Edward~J Hu, Yelong Shen, Phillip Wallis, Zeyuan Allen-Zhu, Yuanzhi Li, Shean
  Wang, Lu~Wang, and Weizhu Chen.
\newblock Lora: Low-rank adaptation of large language models.
\newblock \emph{arXiv preprint arXiv:2106.09685}, 2021.

\bibitem[Immer et~al.(2021)Immer, Korzepa, and Bauer]{immer2021improving}
Alexander Immer, Maciej Korzepa, and Matthias Bauer.
\newblock Improving predictions of bayesian neural nets via local
  linearization.
\newblock In \emph{AISTAT}, 2021.

\bibitem[Izmailov et~al.(2021)Izmailov, Vikram, Hoffman, and
  Wilson]{izmailov2021bayesian}
Pavel Izmailov, Sharad Vikram, Matthew~D Hoffman, and Andrew Gordon~Gordon
  Wilson.
\newblock What are bayesian neural network posteriors really like?
\newblock In \emph{International conference on machine learning}, pp.\
  4629--4640. PMLR, 2021.

\bibitem[Jiang et~al.(2023)Jiang, Sablayrolles, Mensch, Bamford, Chaplot,
  Casas, Bressand, Lengyel, Lample, Saulnier, et~al.]{jiang2023mistral}
Albert~Q Jiang, Alexandre Sablayrolles, Arthur Mensch, Chris Bamford,
  Devendra~Singh Chaplot, Diego de~las Casas, Florian Bressand, Gianna Lengyel,
  Guillaume Lample, Lucile Saulnier, et~al.
\newblock Mistral 7b.
\newblock \emph{arXiv preprint arXiv:2310.06825}, 2023.

\bibitem[Jiang et~al.(2021)Jiang, Araki, Ding, and Neubig]{jiang2021can}
Zhengbao Jiang, Jun Araki, Haibo Ding, and Graham Neubig.
\newblock How can we know when language models know? on the calibration of
  language models for question answering.
\newblock \emph{Transactions of the Association for Computational Linguistics},
  9:\penalty0 962--977, 2021.

\bibitem[Kadavath et~al.(2022)Kadavath, Conerly, Askell, Henighan, Drain,
  Perez, Schiefer, Hatfield-Dodds, DasSarma, Tran-Johnson,
  et~al.]{kadavath2022language}
Saurav Kadavath, Tom Conerly, Amanda Askell, Tom Henighan, Dawn Drain, Ethan
  Perez, Nicholas Schiefer, Zac Hatfield-Dodds, Nova DasSarma, Eli
  Tran-Johnson, et~al.
\newblock Language models (mostly) know what they know.
\newblock \emph{arXiv preprint arXiv:2207.05221}, 2022.

\bibitem[Kristiadi et~al.(2020)Kristiadi, Hein, and Hennig]{kristiadi2020being}
Agustinus Kristiadi, Matthias Hein, and Philipp Hennig.
\newblock Being {B}ayesian, even just a bit, fixes overconfidence in relu
  networks.
\newblock In \emph{ICML}, 2020.

\bibitem[Kunstner et~al.(2019)Kunstner, Hennig, and
  Balles]{kunstner2019limitations}
Frederik Kunstner, Philipp Hennig, and Lukas Balles.
\newblock Limitations of the empirical fisher approximation for natural
  gradient descent.
\newblock \emph{Advances in neural information processing systems}, 32, 2019.

\bibitem[Lakshminarayanan et~al.(2017)Lakshminarayanan, Pritzel, and
  Blundell]{lakshminarayanan2017simple}
Balaji Lakshminarayanan, Alexander Pritzel, and Charles Blundell.
\newblock Simple and scalable predictive uncertainty estimation using deep
  ensembles.
\newblock \emph{Advances in neural information processing systems}, 30, 2017.

\bibitem[Lampinen et~al.(2023)Lampinen, Chan, Dasgupta, Nam, and
  Wang]{lampinenPassiveLearningActive2023}
Andrew~Kyle Lampinen, Stephanie C.~Y. Chan, Ishita Dasgupta, Andrew~J. Nam, and
  Jane~X. Wang.
\newblock Passive learning of active causal strategies in agents and language
  models, May 2023.

\bibitem[Lee et~al.(2019)Lee, Cho, and Kang]{lee2019mixout}
Cheolhyoung Lee, Kyunghyun Cho, and Wanmo Kang.
\newblock Mixout: Effective regularization to finetune large-scale pretrained
  language models.
\newblock \emph{arXiv preprint arXiv:1909.11299}, 2019.

\bibitem[Li et~al.(2022)Li, Puig, Paxton, Du, Wang, Fan, Chen, Huang,
  Aky{\"u}rek, Anandkumar, et~al.]{li2022pre}
Shuang Li, Xavier Puig, Chris Paxton, Yilun Du, Clinton Wang, Linxi Fan, Tao
  Chen, De-An Huang, Ekin Aky{\"u}rek, Anima Anandkumar, et~al.
\newblock Pre-trained language models for interactive decision-making.
\newblock \emph{Advances in Neural Information Processing Systems},
  35:\penalty0 31199--31212, 2022.

\bibitem[Liu et~al.(2022)Liu, Tam, Muqeeth, Mohta, Huang, Bansal, and
  Raffel]{liu2022few}
Haokun Liu, Derek Tam, Mohammed Muqeeth, Jay Mohta, Tenghao Huang, Mohit
  Bansal, and Colin~A Raffel.
\newblock Few-shot parameter-efficient fine-tuning is better and cheaper than
  in-context learning.
\newblock \emph{Advances in Neural Information Processing Systems},
  35:\penalty0 1950--1965, 2022.

\bibitem[Liu et~al.(2019)Liu, Ott, Goyal, Du, Joshi, Chen, Levy, Lewis,
  Zettlemoyer, and Stoyanov]{liu2019roberta}
Yinhan Liu, Myle Ott, Naman Goyal, Jingfei Du, Mandar Joshi, Danqi Chen, Omer
  Levy, Mike Lewis, Luke Zettlemoyer, and Veselin Stoyanov.
\newblock Roberta: A robustly optimized bert pretraining approach.
\newblock \emph{arXiv preprint arXiv:1907.11692}, 2019.

\bibitem[Lu et~al.(2020)Lu, Ie, and Sha]{Lu2020UncertaintyEW}
Zhiyun Lu, Eugene Ie, and Fei Sha.
\newblock Uncertainty estimation with infinitesimal jackknife, its distribution
  and mean-field approximation.
\newblock \emph{ArXiv}, abs/2006.07584, 2020.

\bibitem[MacKay(1998)]{mackay1998choice}
David J.~C. MacKay.
\newblock Choice of basis for laplace approximation.
\newblock \emph{Machine Learning}, 33\penalty0 (1):\penalty0 77--86, 1998.

\bibitem[MacKay(1992)]{mackay1992practical}
David~JC MacKay.
\newblock A practical bayesian framework for backpropagation networks.
\newblock \emph{Neural computation}, 1992.

\bibitem[Mangrulkar et~al.(2022)Mangrulkar, Gugger, Debut, Belkada, and
  Paul]{peft}
Sourab Mangrulkar, Sylvain Gugger, Lysandre Debut, Younes Belkada, and Sayak
  Paul.
\newblock Peft: State-of-the-art parameter-efficient fine-tuning methods.
\newblock \url{https://github.com/huggingface/peft}, 2022.

\bibitem[Mihaylov et~al.(2018)Mihaylov, Clark, Khot, and
  Sabharwal]{OpenBookQA2018}
Todor Mihaylov, Peter Clark, Tushar Khot, and Ashish Sabharwal.
\newblock Can a suit of armor conduct electricity? a new dataset for open book
  question answering.
\newblock In \emph{EMNLP}, 2018.

\bibitem[Ober \& Aitchison(2021)Ober and Aitchison]{ober2021global}
Sebastian~W Ober and Laurence Aitchison.
\newblock Global inducing point variational posteriors for bayesian neural
  networks and deep gaussian processes.
\newblock In \emph{International Conference on Machine Learning}, pp.\
  8248--8259. PMLR, 2021.

\bibitem[OpenAI(2023)]{openai2023gpt4}
OpenAI.
\newblock {GPT}-4 technical report, 2023.

\bibitem[Osawa et~al.(2023)Osawa, Ishikawa, Yokota, Li, and
  Hoefler]{osawa2023asdl}
Kazuki Osawa, Satoki Ishikawa, Rio Yokota, Shigang Li, and Torsten Hoefler.
\newblock Asdl: A unified interface for gradient preconditioning in pytorch.
\newblock \emph{arXiv preprint arXiv:2305.04684}, 2023.

\bibitem[Ouyang et~al.(2022)Ouyang, Wu, Jiang, Almeida, Wainwright, Mishkin,
  Zhang, Agarwal, Slama, Ray, et~al.]{ouyang2022training}
Long Ouyang, Jeffrey Wu, Xu~Jiang, Diogo Almeida, Carroll Wainwright, Pamela
  Mishkin, Chong Zhang, Sandhini Agarwal, Katarina Slama, Alex Ray, et~al.
\newblock Training language models to follow instructions with human feedback.
\newblock \emph{Advances in Neural Information Processing Systems},
  35:\penalty0 27730--27744, 2022.

\bibitem[Park \& Caragea(2022)Park and Caragea]{park2022calibration}
Seo~Yeon Park and Cornelia Caragea.
\newblock On the calibration of pre-trained language models using mixup guided
  by area under the margin and saliency.
\newblock \emph{arXiv preprint arXiv:2203.07559}, 2022.

\bibitem[Ritter et~al.(2018)Ritter, Botev, and Barber]{ritter2018scalable}
Hippolyt Ritter, Aleksandar Botev, and David Barber.
\newblock A scalable laplace approximation for neural networks.
\newblock In \emph{ICLR}, 2018.

\bibitem[Sakaguchi et~al.(2021)Sakaguchi, Bras, Bhagavatula, and
  Choi]{sakaguchi2021winogrande}
Keisuke Sakaguchi, Ronan~Le Bras, Chandra Bhagavatula, and Yejin Choi.
\newblock Winogrande: An adversarial winograd schema challenge at scale.
\newblock \emph{Communications of the ACM}, 2021.

\bibitem[Shi \& Lipani(2023)Shi and Lipani]{shi2023dept}
Zhengxiang Shi and Aldo Lipani.
\newblock Dept: Decomposed prompt tuning for parameter-efficient fine-tuning.
\newblock \emph{arXiv preprint arXiv:2309.05173}, 2023.

\bibitem[Shridhar et~al.(2019)Shridhar, Laumann, and
  Liwicki]{shridhar2019comprehensive}
Kumar Shridhar, Felix Laumann, and Marcus Liwicki.
\newblock A comprehensive guide to bayesian convolutional neural network with
  variational inference.
\newblock \emph{arXiv preprint arXiv:1901.02731}, 2019.

\bibitem[Singhal et~al.(2022)Singhal, Azizi, Tu, Mahdavi, Wei, Chung, Scales,
  Tanwani, {Cole-Lewis}, Pfohl, Payne, Seneviratne, Gamble, Kelly, Scharli,
  Chowdhery, Mansfield, y~Arcas, Webster, Corrado, Matias, Chou, Gottweis,
  Tomasev, Liu, Rajkomar, Barral, Semturs, Karthikesalingam, and
  Natarajan]{singhalLargeLanguageModels2022}
Karan Singhal, Shekoofeh Azizi, Tao Tu, S.~Sara Mahdavi, Jason Wei, Hyung~Won
  Chung, Nathan Scales, Ajay Tanwani, Heather {Cole-Lewis}, Stephen Pfohl,
  Perry Payne, Martin Seneviratne, Paul Gamble, Chris Kelly, Nathaneal Scharli,
  Aakanksha Chowdhery, Philip Mansfield, Blaise~Aguera y~Arcas, Dale Webster,
  Greg~S. Corrado, Yossi Matias, Katherine Chou, Juraj Gottweis, Nenad Tomasev,
  Yun Liu, Alvin Rajkomar, Joelle Barral, Christopher Semturs, Alan
  Karthikesalingam, and Vivek Natarajan.
\newblock Large {{Language Models Encode Clinical Knowledge}}.
\newblock \emph{{arXiv}}, 2022.

\bibitem[Tian et~al.(2023)Tian, Mitchell, Zhou, Sharma, Rafailov, Yao, Finn,
  and Manning]{tian2023just}
Katherine Tian, Eric Mitchell, Allan Zhou, Archit Sharma, Rafael Rafailov,
  Huaxiu Yao, Chelsea Finn, and Christopher~D Manning.
\newblock Just ask for calibration: Strategies for eliciting calibrated
  confidence scores from language models fine-tuned with human feedback.
\newblock \emph{arXiv preprint arXiv:2305.14975}, 2023.

\bibitem[Touvron et~al.(2023{\natexlab{a}})Touvron, Lavril, Izacard, Martinet,
  Lachaux, Lacroix, Rozi{\`e}re, Goyal, Hambro, Azhar,
  et~al.]{touvron2023llama}
Hugo Touvron, Thibaut Lavril, Gautier Izacard, Xavier Martinet, Marie-Anne
  Lachaux, Timoth{\'e}e Lacroix, Baptiste Rozi{\`e}re, Naman Goyal, Eric
  Hambro, Faisal Azhar, et~al.
\newblock Llama: Open and efficient foundation language models.
\newblock \emph{arXiv preprint arXiv:2302.13971}, 2023{\natexlab{a}}.

\bibitem[Touvron et~al.(2023{\natexlab{b}})Touvron, Martin, Stone, Albert,
  Almahairi, Babaei, Bashlykov, Batra, Bhargava, Bhosale,
  et~al.]{touvron2023llama2}
Hugo Touvron, Louis Martin, Kevin Stone, Peter Albert, Amjad Almahairi, Yasmine
  Babaei, Nikolay Bashlykov, Soumya Batra, Prajjwal Bhargava, Shruti Bhosale,
  et~al.
\newblock Llama 2: Open foundation and fine-tuned chat models.
\newblock \emph{arXiv preprint arXiv:2307.09288}, 2023{\natexlab{b}}.

\bibitem[Tran et~al.(2019)Tran, Dusenberry, Van Der~Wilk, and
  Hafner]{tran2019bayesian}
Dustin Tran, Mike Dusenberry, Mark Van Der~Wilk, and Danijar Hafner.
\newblock Bayesian layers: A module for neural network uncertainty.
\newblock \emph{Advances in neural information processing systems}, 32, 2019.

\bibitem[Wang et~al.(2019{\natexlab{a}})Wang, Pruksachatkun, Nangia, Singh,
  Michael, Hill, Levy, and Bowman]{wang2019superglue}
Alex Wang, Yada Pruksachatkun, Nikita Nangia, Amanpreet Singh, Julian Michael,
  Felix Hill, Omer Levy, and Samuel Bowman.
\newblock Superglue: A stickier benchmark for general-purpose language
  understanding systems.
\newblock In \emph{NeurIPS}, 2019{\natexlab{a}}.

\bibitem[Wang et~al.(2019{\natexlab{b}})Wang, Singh, Michael, Hill, Levy, and
  Bowman]{wang2018glue}
Alex Wang, Amanpreet Singh, Julian Michael, Felix Hill, Omer Levy, and Samuel~R
  Bowman.
\newblock Glue: A multi-task benchmark and analysis platform for natural
  language understanding.
\newblock In \emph{ICLR}, 2019{\natexlab{b}}.

\bibitem[Wang et~al.(2023)Wang, Aitchison, and Rudolph]{wang2023lora}
Xi~Wang, Laurence Aitchison, and Maja Rudolph.
\newblock Lora ensembles for large language model fine-tuning.
\newblock \emph{arXiv preprint arXiv:2310.00035}, 2023.

\bibitem[Wang et~al.(2022)Wang, Kordi, Mishra, Liu, Smith, Khashabi, and
  Hajishirzi]{wang2022self}
Yizhong Wang, Yeganeh Kordi, Swaroop Mishra, Alisa Liu, Noah~A Smith, Daniel
  Khashabi, and Hannaneh Hajishirzi.
\newblock Self-instruct: Aligning language model with self generated
  instructions.
\newblock \emph{arXiv preprint arXiv:2212.10560}, 2022.

\bibitem[Wei et~al.(2021)Wei, Bosma, Zhao, Guu, Yu, Lester, Du, Dai, and
  Le]{wei2021finetuned}
Jason Wei, Maarten Bosma, Vincent~Y Zhao, Kelvin Guu, Adams~Wei Yu, Brian
  Lester, Nan Du, Andrew~M Dai, and Quoc~V Le.
\newblock Finetuned language models are zero-shot learners.
\newblock \emph{arXiv preprint arXiv:2109.01652}, 2021.

\bibitem[Wolf et~al.(2020)Wolf, Debut, Sanh, Chaumond, Delangue, Moi, Cistac,
  Rault, Louf, Funtowicz, Davison, Shleifer, von Platen, Ma, Jernite, Plu, Xu,
  Scao, Gugger, Drame, Lhoest, and Rush]{wolf2020transformers}
Thomas Wolf, Lysandre Debut, Victor Sanh, Julien Chaumond, Clement Delangue,
  Anthony Moi, Pierric Cistac, Tim Rault, Rémi Louf, Morgan Funtowicz, Joe
  Davison, Sam Shleifer, Patrick von Platen, Clara Ma, Yacine Jernite, Julien
  Plu, Canwen Xu, Teven~Le Scao, Sylvain Gugger, Mariama Drame, Quentin Lhoest,
  and Alexander~M. Rush.
\newblock Transformers: State-of-the-art natural language processing.
\newblock In \emph{EMNLP}, 2020.

\bibitem[Wu et~al.(2023)Wu, Irsoy, Lu, Dabravolski, Dredze, Gehrmann, Kambadur,
  Rosenberg, and Mann]{wuBloombergGPTLargeLanguage2023}
Shijie Wu, Ozan Irsoy, Steven Lu, Vadim Dabravolski, Mark Dredze, Sebastian
  Gehrmann, Prabhanjan Kambadur, David Rosenberg, and Gideon Mann.
\newblock {{BloombergGPT}}: {{A Large Language Model}} for {{Finance}}, May
  2023.

\bibitem[Xiao et~al.(2022)Xiao, Liang, Bhatt, Neiswanger, Salakhutdinov, and
  Morency]{xiao2022uncertainty}
Yuxin Xiao, Paul~Pu Liang, Umang Bhatt, Willie Neiswanger, Ruslan
  Salakhutdinov, and Louis-Philippe Morency.
\newblock Uncertainty quantification with pre-trained language models: A
  large-scale empirical analysis.
\newblock In \emph{EMNLP}, 2022.

\bibitem[Xue et~al.(2021)Xue, Yu, Xu, Liu, Hu, Ye, Geng, Liu, and
  Meng]{xue2021bayesian}
Boyang Xue, Jianwei Yu, Junhao Xu, Shansong Liu, Shoukang Hu, Zi~Ye, Mengzhe
  Geng, Xunying Liu, and Helen Meng.
\newblock Bayesian transformer language models for speech recognition.
\newblock In \emph{ICASSP 2021-2021 IEEE International Conference on Acoustics,
  Speech and Signal Processing (ICASSP)}, pp.\  7378--7382. IEEE, 2021.

\bibitem[Zhai et~al.(2023)Zhai, Zhang, Lei, Yu, Xu, Feng, Ding, and
  Wang]{zhai2023uncertainty}
Yuanzhao Zhai, Han Zhang, Yu~Lei, Yue Yu, Kele Xu, Dawei Feng, Bo~Ding, and
  Huaimin Wang.
\newblock Uncertainty-penalized reinforcement learning from human feedback with
  diverse reward lora ensembles.
\newblock \emph{arXiv preprint arXiv:2401.00243}, 2023.

\bibitem[Zhang et~al.(2017)Zhang, Cisse, Dauphin, and
  Lopez-Paz]{zhang2017mixup}
Hongyi Zhang, Moustapha Cisse, Yann~N Dauphin, and David Lopez-Paz.
\newblock mixup: Beyond empirical risk minimization.
\newblock \emph{arXiv preprint arXiv:1710.09412}, 2017.

\bibitem[Zhang et~al.(2019)Zhang, Li, Zhang, Chen, and
  Wilson]{zhang2019cyclical}
Ruqi Zhang, Chunyuan Li, Jianyi Zhang, Changyou Chen, and Andrew~Gordon Wilson.
\newblock Cyclical stochastic gradient mcmc for bayesian deep learning.
\newblock \emph{arXiv preprint arXiv:1902.03932}, 2019.

\bibitem[Zhang et~al.(2021)Zhang, Fan, Chen, and Zhou]{zhang2021bayesian}
Shujian Zhang, Xinjie Fan, Bo~Chen, and Mingyuan Zhou.
\newblock Bayesian attention belief networks.
\newblock In \emph{ICML}, 2021.

\bibitem[Zhao et~al.(2020)Zhao, Chen, and Oymak]{zhao2020role}
Yuan Zhao, Jiasi Chen, and Samet Oymak.
\newblock On the role of dataset quality and heterogeneity in model confidence.
\newblock \emph{arXiv preprint arXiv:2002.09831}, 2020.

\end{thebibliography}
\bibliographystyle{iclr2024_conference}

\appendix

\newpage

\section{Laplace predictive posterior approximations} \label{app:laplace_approx}
In this section, we review several methods for obtaining the predictive posterior of Laplace approximation.

\subsection{Monte Carlo sampling}
As descibed in main text, we can obtain a closed-form Gaussian posterior on output logits,
\begin{align}
    f_\param(\x_*) \sim \N{f_{\param_\text{MAP}}(\x_*),\mL},
\end{align}
where
\begin{align}
    \mL = (\nabla_\param f_{\param}(\x_*)|^T_{\param_\text{MAP}}) \S (\nabla_\param f_{\param}(\x_*)|_{\param_\text{MAP}}).
\end{align}
To obtain samples of $f_\param(\x_*)$, we can decompose the covariance using the Cholesky factorization $\mL =  \Lm\Lm^T$ with
\begin{align}
    \tilde{f}_\param(\x_*) = f_{\param_\text{MAP}}(\x_*) + \Lm \boldsymbol{\xi},
\end{align}
where $\boldsymbol{\xi}$ is a vector of IID standard normal random variables. We can compute the Bayesian model average by computing the average probabilities (passing the sampled logits through softmax function) under the Gaussian random noise from $\boldsymbol{\xi}$.

\subsection{Probit approximation}
A closed-form approximation of the predictive posterior for classification can be obtained by integrating out the posterior over weights with a generalized probit approximation \citep{Lu2020UncertaintyEW,daxberger2021laplace} of the likelihood,
\begin{align} \label{eq:pred}
    p(y_*|\x_*,\mathcal{D}) \approx \text{Categorical}\b{y_*, \softmax\b{\frac{f_{\param_\text{MAP}}(\x_*)}{\sqrt{1+\tfrac{\pi}{8}\diag(\mL)}}} },
\end{align}
where
\begin{align}
    \mL = (\nabla_\param f_{\param}(\x_*)|^T_{\param_\text{MAP}}) \S (\nabla_\param f_{\param}(\x_*)|_{\param_\text{MAP}}).
\end{align}
Although probit approximation provably preserves decision boundary in binary sigmoid classification \citep{kristiadi2020being}, it does hold for softmax multiclass classification.

\subsection{Laplace bridge}
The Laplace bridge \citep{mackay1998choice,daxberger2021laplace} maps a Gaussian $\mathcal{N}(\bm{\mu},\S)$ to a Dirichlet distribution $\mathcal{D}(\bm{\alpha})$ over classes with
\begin{align}
    \bm{\alpha}_i = \frac{1}{\S_{ii}} \bigg( 1 - \frac{2}{C} + \frac{\exp(\bm{\mu}_i)}{C^2} \sum_j \exp(-\bm{\mu}_j) \bigg),
\end{align}
where $C$ denotes the number of classes. Similar to the generalized probit approximation, it also ignores the covariance terms and only considers the diagonal of the covariance $\S_{ii}$.

\subsection{Empirical comparison}
\begin{figure*}[t]
    \centering
    \includegraphics[width=\textwidth]{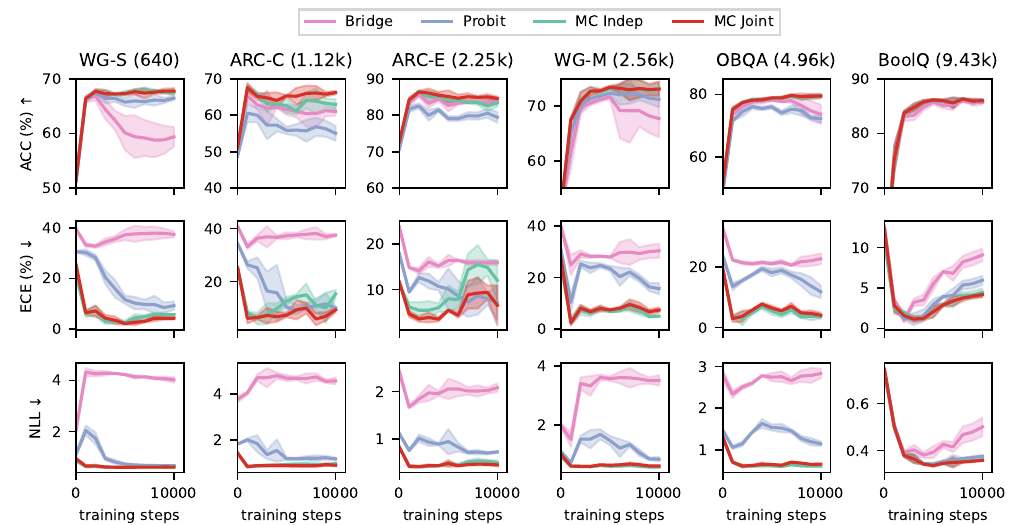}
    \caption{Fine-tuning of LlaMA2-7B across six common sense reasoning tasks, comparing different Laplace predictive posterior approximations: Laplace bridge approximation (bridge), generalized probit approximation (probit), Monte Carlo sampling using the diagonal covariance (MC indep), and Monte Carlo sampling using the full covariance (MC joint).}
    \label{fig:laplace_approx}
\end{figure*}

Figure~\ref{fig:laplace_approx} shows a comparison of these approximations in fine-tuning Llama2-7B and applying full Laplace-LoRA post-hoc. Specifically, we considered Monte Carlo sampling using the full covariance (MC joint), Monte Carlo sampling only using the diagonal covariance (MC indep), generalized probit approximation (probit), and Laplace bridge (bridge).
MC joint consistently achieves highest accuracy and among the best NLL, while bridge is often the worst, probit and MC indep can sometimes give suboptimal performance. This is likely due to bridge, probit and MC indep are all approximations that ignored the covariances between logits, whereas MC joint faithfully approximates the true predictive posterior.

\section{Prompt templates for LlaMA2 common sense reasoning tasks} \label{app:prompt}
We present our prompt templates used to fine-tune LlaMA2 on common sense reasoning tasks in Table~\ref{table:llama_prompt}. For ARC datasets, although the majority of quesstions have four choices, there are a tiny amount of questions with three or five choices which we remove for consistency. In ARC-C, there are 1/1119 with three choices and 1/1119 with five choices in the training set; 3/299 with three choices and 1/299 with five choices in the evaluation set. In ARC-E, there are 6/2251 with three choices and 4/2251 with five choices in the trainng set; 1/570 with three choices and 2/570 with five choices in the evaluation set.

\begin{table}[t]
\centering
\captionsetup{skip=8pt}
\begin{tabular}{c|c}
\hline
\textbf{Task} & \textbf{Prompt} \\
\hline
Winogrande (WG-S/WG-M) & Select one of the choices that answers the following question: \\
& \{\texttt{question}\} Choices: A. \{\texttt{option1}\}. B \{\texttt{option2}\}. Answer: \\
\hline
ARC (ARC-C/ARC-E) & Select one of the choices that answers the following question: \\
& \{\texttt{question}\} Choices: A. \{\texttt{choice1}\}. B. \{\texttt{choice2}\}. \\
& C. \{\texttt{choice2}\}. D. \{\texttt{choice2}\}. Answer: \\
\hline
Openbook QA (OBQA) & Select one of the choices that answers the following question: \\
& \{\texttt{question}\} Choices: A. \{\texttt{choice1}\}. B. \{\texttt{choice2}\}. \\
& C. \{\texttt{choice2}\}. D. \{\texttt{choice2}\}. Answer: \\
\hline
BoolQ & Answer the question with only True or False: \\
& \{\texttt{question}\} Context: \{\texttt{passage}\}. \\
\hline
\end{tabular}
\caption{Prompt templates for fine-tuning LlaMA-7B on common sense reasoning tasks.}
\label{table:llama_prompt}
\end{table}

\section{Hyperparameters} \label{app:hyperparam}
We follow the default hyperparameters from Huggingface's Transformer \citep{wolf2020transformers} and PEFT \citep{peft} libraries. Hyperparameters used for fine-tuning RoBERTa-base and RoBERTa-large with LoRA are shown in Table~\ref{table:roberta_hyper}, those for fine-tuning LlaMA-7B are shown in Table~\ref{table:llama_hyper}. Note the only differences between the two settings are a smaller batch size to fit into our GPU memory, and a longer max sequence length to account for prompt length.

\begin{table}[t]
\centering
\captionsetup{skip=8pt}
\begin{tabular}{c|c}
\hline
\textbf{Hyperparameter} & \textbf{Value} \\
\hline
LoRA \( r \) & 8 \\
LoRA \( \alpha \) & 16 \\
Dropout Probability & 0.1 \\
Weight Decay & 0 \\
Learning Rate & \( 5 \times 10^{-5} \) \\
Learning Rate Scheduler & Linear \\
Batch Size & 32 \\
Max Sequence Length & 256 \\
\hline
\end{tabular}
\caption{Hyperparameters used in fine-tuning RoBERTa-base and RoBERTa-large with LoRA.}
\label{table:roberta_hyper}
\end{table}

\begin{table}[t]
\centering
\captionsetup{skip=8pt}
\begin{tabular}{c|c}
\hline
\textbf{Hyperparameter} & \textbf{Value} \\
\hline
LoRA \( r \) & 8 \\
LoRA \( \alpha \) & 16 \\
Dropout Probability & 0.1 \\
Weight Decay & 0 \\
Learning Rate & \( 5 \times 10^{-5} \) \\
Learning Rate Scheduler & Linear \\
Batch Size & 4 \\
Max Sequence Length & 300 \\
\hline
\end{tabular}
\caption{Hyperparameters used in fine-tuning LlaMA-7B with LoRA.}
\label{table:llama_hyper}
\end{table}


\section{Metrics for uncertainty quantification} \label{app:metrics}
There are two commonly used metrics for measuring uncertainty quantification in neural networks: negative log-likelihood (NLL) and expected calibration error (ECE). NLL computes the sum of negative expected log probability of predicting the true label,
\begin{align}
    \text{NLL} =  \sum_{i=1}^N -\log P(\hat{y}_i = y_i),
\end{align}
where $P(\hat{y}_i)$ is the model's output distribution, $y_i$ is the true label. NLL is also equivalent to cross-entropy between the one-hot data distribution and the model's output distribution. NLL encourages the model to give high probability to correct answers. If the model is overconfident in a wrong answer, then the probability of giving the right answer would be low which raises NLL. 
On the other hand, ECE measures the alignment between model's confidence and accuracy, by binning the highest predicted probabilities and compute a weighted average between the difference of average accuracy and confidence in each bin,
\begin{align}
    \text{ECE} = \sum_{m=1}^M \frac{|B_m|}{N} |\text{acc}(B_m) - \text{conf}(B_m)|,
\end{align}
where $\text{acc}(B_m)$ and $\text{conf}(B_m)$ are the average accuracy and average confidence in each bin, 
\begin{align}
    \text{acc}(B_m) = \frac{1}{|B_m|} \sum_{i\in B_m} \mathbf{1}(\hat{y}_i = y_i), \qquad \text{conf}(B_m) = \frac{1}{|B_m|} \sum_{i\in B_m} P(\hat{y}_i),
\end{align}
and $|B_m|$ is the number of examples in bin $m$. However, expected calibration error cannot be optimized directly like negative log-likelihood, as a completely random model will have the same accuracy and confidence for each datapoint, thus achieving zero ECE \citep{ashukha2020pitfalls}.

\section{Low-rank computations}
\label{app:low-rank}
The usual K-FAC approach is to write,
\begin{align}
  \Slike &= \SlikeA \otimes \SlikeB 
  \intertext{As this is for the LoRA adapters, there's one small and one large covariance matrix.
  Without loss of generality, assume that $\SlikeA \in \mathbb{R}^{\nlora \times \nlora}$ is the small matrix, so we represent it as full-rank, and we assume that $\SlikeB \in \mathbb{R}^{d \times d}$ is the large matrix, so we represent it as a low-rank,}
  \SlikeB^{-1} &\approx \B \B^T.
\end{align}
Thus, we write $\SlikeB$ in low-rank form, in terms of $\B \in \mathbb{R}^{d \times \nkfac}$.

The posterior precision is the sum of the likelihood and prior precisions,
\begin{align}
  \Spost^{-1} &= \SlikeA^{-1} \otimes \SlikeB^{-1} + \tfrac{1}{\sigma^2} \I \otimes \I.\\
  \intertext{Substituting our low-rank approximation for the large Kronecker factor,}
  \Spost^{-1} &= \SlikeA^{-1} \otimes \b{\B \B^T} + \tfrac{1}{\sigma^2}\I \otimes \I.\\
  \intertext{Using the mixed product property, and setting $\SlikeA^{-1} = \mathbf{L} \mathbf{L}^T$ (e.g. $\mathbf{L}$ is the Cholesky decomposition of $\Slike^{-1}$),} 
  \Spost^{-1} &= \b{\mathbf{L} \otimes \B} \b{\mathbf{L} \otimes \B}^T + \tfrac{1}{\sigma^2}\I \otimes \I.
\end{align}
This is in the form of a diagonal matrix, $\tfrac{1}{\sigma^2}\I \otimes \I$ plus a low-rank matrix, $\b{\mathbf{L} \otimes \B} \b{\mathbf{L} \otimes \B}^T$.
Note that the rank of the low-rank matrix here is $\nlora \nkfac$.
We're going to compute everything we need using this low-rank form.

\subsection{Incremental computation of low-rank factor}
\label{app:lr_incremental}

The first question is how to estimate the low-rank components, without ever computing a large $d \times d$ matrix.
The answer --- an incremental low-rank SVD --- is given in Alg.~\ref{alg:incremental}.
\begin{algorithm}[H]   
  \caption{Memory efficient estimate of low-rank $\B$ such that $\B \B^T \approx \sum_{t=1}^T \bv_t \bv_t^T$. \label{alg:incremental}}
  \begin{algorithmic}
    \STATE \textcolor{gray}{Initialize to $0$}
    \STATE $\B = \0$ 
    \FOR{$t = 1,...,T$}
    \STATE \textcolor{gray}{Combine current low-rank estimate, $\B$, with new activities, $\bv_t=\a_t$, or new gradients, $\bv_t=\g_t$}
    \STATE \textcolor{gray}{Could combine with several vectors, if available}
    \STATE $\B' = \begin{pmatrix} \B & \bv_t \end{pmatrix}$
    \STATE \textcolor{gray}{New low-rank estimate from top $\nkfac$ components of SVD}
    \STATE \textcolor{gray}{$\V$ doesn't matter as its just a rotation}
    \STATE $\U, \mathbf{S}, \V^T = \text{svd}(\B')$
    \STATE $\B \leftarrow \U[:, :\nkfac] \mathbf{S}[:\nkfac, :\nkfac]$
    \ENDFOR
  \end{algorithmic}
\end{algorithm}

\subsection{Marginal likelihood optimization with the matrix determinant lemma}
\label{app:lr_marg_like}

To optimize the marginal likelihood (Eq.), we need the log-determinant of the posterior.
To compute this log-determinant efficient in the low-rank setting, we use the matrix determinant lemma,
\begin{align}
  \log \det\b{\Spost^{-1}} &= \log \det\b{\tfrac{1}{\sigma^2}\I \otimes \I} + \log \det\b{\I + \sigma^2 \b{\mathbf{L} \otimes \B}^T \b{\mathbf{L} \otimes \B}}
  \intertext{Applying the mixed product property,}
  \log \det\b{\Spost^{-1}} &= - \nlora d \log \sigma^2 + \log \det\b{\mathbf{M}}
  \intertext{where,}
  \mathbf{M} &= \big(\I_{\nlora \nkfac} + \sigma^2 \underbrace{\underbrace{\mathbf{L}^T \mathbf{L}}_{\nlora \times \nlora} \otimes \underbrace{\B^T \B}_{\nkfac \times \nkfac}}_{\nlora \nkfac \times \nlora \nkfac}\big)
\end{align}
So we can explicitly compute $\mathbf{L}^T \mathbf{L} \otimes \B^T \B$ and hence $\mathbf{M}\in\mathbb{R}^{\nlora \nkfac \times \nlora \nkfac}$ as long as $\nlora \nkfac$ is not too big.
In our case, $\nlora=4$ and $\nkfac=10$, so $\nlora \nkfac=40$.

\subsection{Linearized prediction with Woodbury}
\label{app:lr_linearised}

For prediction, we need to use Woodbury to compute:
\begin{align}
  \Spost &= \b{\b{\mathbf{L} \otimes \B} \b{\mathbf{L} \otimes \B}^T + \tfrac{1}{\sigma^2}\I \otimes \I}^{-1}\\
  \Spost &= \sigma^2 \I_{\nlora} \otimes \I_{d} - \sigma^4 \b{\mathbf{L} \otimes \B} \mathbf{M}^{-1} \b{\mathbf{L} \otimes \B}^T 
\end{align}
Now, applying this form for prediction,
\begin{align}
  \Lambda_{ij} &= \g_i^T \Spost \g_j.
  \intertext{where $\g_i$ is the gradient of the $i$th network output for a particular parameter, i.e.\ part of the Jacobian,}
  \g_i &= \frac{\partial f_i}{\partial \boldsymbol{\theta}} 
\end{align}
This is really a sum over layers, so we now just think of a single layer.
Thus, $\g_i \in \mathbb{R}^{\nlora d}$ is really the whole matrix of gradients,
\begin{align}
  \g_i &= \vec\b{\G_i}.
  \intertext{where $\G_i \in \mathbb{R}^{d \times \nlora}$. Thus,}
  \Lambda_{ij} &= \vec\b{\G_i}^T \Spost \vec\b{\G_j}\\
  \Lambda_{ij} &= \vec\b{\G_i}^T \b{\sigma^2 \I_{\nlora} \otimes \I_{d} - \sigma^4 \b{\mathbf{L} \otimes \B} \mathbf{M}^{-1} \b{\mathbf{L} \otimes \B}^T} \vec\b{\G_j}\\
  \Lambda_{ij} &= \sigma^2 \vec\b{\G_i}^T \vec\b{\G_j}-  \sigma^4 \vec\b{\G_i}^T \b{\mathbf{L} \otimes \B} \mathbf{M}^{-1} \b{\mathbf{L}^T \otimes \B^T} \vec\b{\G_j}\\
  \intertext{Using the mixed Kronecker matrix-vector product,}
  \Lambda_{ij} &= \sigma^2 \vec\b{\G_i}^T \vec\b{\G_j}-  \sigma^4 \vec\b{\B^T \G_i \mathbf{L}}^T \mathbf{M}^{-1} \vec\b{\B^T \G_j \mathbf{L}}.
\end{align}

\section{Optimizing Laplace prior precision} \label{app:laplace_prior}
In this section, we present how we optimize the Laplace prior precision. When there is only a training set available with no validation set (such as in Figure~\ref{fig:llama7b-kron}, \ref{fig:roberta-base-kron}, \ref{fig:roberta-large-kron} and Table~\ref{table:id_train}, \ref{table:roberta-base_train}, \ref{table:roberta-large_train}), we can use the Laplace model evidence in Equation~\ref{eq:model_evidence} to optimize prior precision. Our algorithm is presented in Algorithm~\ref{alg:LA_evidence}, and we chose $\eta=0.1$, and $M=100$.

\begin{algorithm}[t]   
    \caption{Optimize Laplace prior precision using the training set model evidence} \label{alg:LA_evidence}
    \begin{algorithmic}
            \STATE Initialize prior precision $\lambda$, learning rate $\eta$, optimization steps $M$
            \STATE Obtain $\theta_\text{MAP}$ from a fine-tuned checkpoint, pre-compute $\log \Pc{\y}{\X, \param}$ and Fisher $\F$
            \FOR{$step = 1,...,M$}
            \STATE
            Compute posterior covariance $\S = \F + \lambda \I$
            \STATE  
            Calculate $\L{\param} = \log \Pc{\y}{\X, \param} + \log \P{\param} = \log \Pc{\y}{\X, \param} + \lambda ||\param_\text{MAP}||_2^2$
            \STATE Perform a gradient step with respect to $\lambda$ to maximize log model evidence (Eq.\ref{eq:model_evidence}): \\
            \qquad $\lambda \leftarrow \lambda + \eta \nabla_{\lambda} \big( \L{\paramMAP} + \tfrac{1}{2}\log|\S| \big)$
            \ENDFOR
    \end{algorithmic}
\end{algorithm}

When we introduce a validation set by splitting the training set (such as in Figure~\ref{fig:llama7b-kron-val}, \ref{fig:roberta-base-kron_val}, \ref{fig:roberta-large-kron_val} and Table~\ref{table:id_val}, \ref{table:roberta_base_id_val}, \ref{table:roberta_large_id_val}), we can use the validation log-likelihood to optimize the Laplace prior precision. For memory and computational efficiency, we precompute the mean $f_{\param_\text{MAP}}$ and the Jacobian $\J = \nabla_\param f_{\param}(\X)|_{\param_\text{MAP}}$, then perform mini-batch gradient descent on $\lambda$ (reparametrization in Bayesian model average allows gradient flowing through) as detailed in Algorithm~\ref{alg:LA_val}. We chose $\eta=0.1$, $M=1000$, and $b=4$.

\begin{algorithm}[t]   
    \caption{Optimize Laplace prior precision using validation log-likelihood} \label{alg:LA_val}
    \begin{algorithmic}
            \STATE Initialize prior precision $\lambda$, learning rate $\eta$, batch size $b$, validation set $(\X,\y)$
            \STATE Obtain $\param_\text{MAP}$ from a fine-tuned checkpoint, pre-compute mean $f_{\param_\text{MAP}}$, Jacobian $\J = \nabla_\param f_{\param}(\X)|_{\param_\text{MAP}}$, Fisher $\F$
            \FOR{$step = 1,...,M$}
            \STATE
            Randomly sample a batch of validation data $\X_b,\y_b$ with corresponding Jacobian $\J_b$
            \STATE
            Compute posterior covariance $\S = \F + \lambda \I$
            \STATE
            Calculate batch logits covariance $\mL_b = \J_b^T \S \J_b$ and Cholesky $\mL_b =  \Lm_b\Lm_b^T$
            \STATE
            Obtain batch Bayesian model average $\tilde{f}_\param(\X_b) = f_{\param_\text{MAP}}(\X_b) + \Lm_b \boldsymbol{\xi}$
            \STATE
            Evaluate validation likelihood $\Pc{\y_b}{\X_b, \param} = \text{Categorical}\big(\y_b; \softmax(\tilde{f}_\param(\X_b)) \big)$
            \STATE
            Perform a gradient step  with respect to $\lambda$ to maximize mini-batch validation log-likelihood: \\
            \qquad $\lambda \leftarrow \lambda + \eta \nabla_{\lambda} \log \Pc{\y_b}{\X_b, \param}$
            \ENDFOR
    \end{algorithmic}
\end{algorithm}


\section{Additional experiments}

\subsection{LlaMA2-7B evaluations under distribution shift}
Table~\ref{table:ood_val} shows the {LlaMA2-7B evaluations under distribution shift results using the best validation accuracy checkpoint on OBQA.

\begin{table}[t] 
\small
\centering
\caption{Comparison of different post-hoc methods across six out-of-distribution datasets. Results are evaluated using the fine-tuned LlaMA2-7B on OBQA as in Table~\ref{table:id_val}.}
\label{table:ood_val}
\vspace{0.1cm}
\resizebox{0.975\textwidth}{!}{\begin{tabular}{l|| l | l |l l | l l l l}

&  & \multicolumn{1}{c|}{ID} & \multicolumn{2}{c|}{Smaller Distribution Shift} & \multicolumn{4}{c}{Larger Distribution Shift} \\
\rule{0pt}{2.25ex}
Metrics & Methods & OBQA & ARC-C & ARC-E & CS & Eng & Law & Health \\
\hline\hline
\multirow{6}{*}{ACC $\uparrow$} \rule{0pt}{2.25ex}
& MAP    & $77.7_{0.8}$ & $68.0_{0.2}$ & $76.7_{1.0}$ & $43.5_{0.9}$ & $44.4_{2.0}$ & $37.4_{0.1}$ & $47.7_{0.8}$ \\
& MC Drop     & $77.7_{0.2}$ & $68.7_{0.2}$ & $77.2_{1.2}$ & $42.2_{1.3}$ & $45.6_{1.6}$ & $37.2_{0.2}$ & $47.9_{0.7}$ \\
& Ckpt Ens     & $78.2_{0.2}$ & $67.9_{0.2}$ & $77.8_{0.4}$ & $42.4_{0.7}$ & $43.5_{0.9}$ & $37.1_{0.4}$ & $48.1_{0.2}$ \\
& Temp   & $77.7_{0.8}$ & $68.0_{0.2}$ & $76.7_{1.0}$ & $43.5_{0.9}$ & $44.4_{2.0}$ & $37.4_{0.1}$ & $47.7_{0.8}$ \\
\cdashline{2-9} \rule{0pt}{2.25ex}
& LLLA   & $77.6_{0.7}$ & $67.8_{0.0}$ & $77.4_{0.0}$ & $42.9_{0.0}$ & $43.8_{0.0}$ & $36.9_{0.0}$ & $46.6_{0.0}$ \\
& LA     & $78.1_{0.7}$ & $67.8_{0.0}$ & $76.7_{0.0}$ & $44.1_{0.0}$ & $45.8_{0.0}$ & $37.2_{0.0}$ & $46.6_{0.0}$ \\
\hline
\multirow{6}{*}{ECE $\downarrow$} \rule{0pt}{2.25ex}
& MAP    & $9.8_{1.0}$ & $12.2_{0.7}$ & $8.6_{1.5}$ & $22.0_{1.5}$ & $20.7_{1.0}$ & $28.4_{0.8}$ & $23.2_{0.7}$ \\
& MC Drop     & $8.8_{0.8}$ & $12.9_{0.8}$ & $7.7_{1.8}$ & $22.8_{2.0}$ & $19.6_{1.5}$ & $28.5_{0.6}$ & $22.6_{0.6}$ \\
& Ckpt Ens     & $4.7_{0.5}$ & $7.1_{1.0}$ & $3.5_{0.3}$ & $17.4_{0.4}$ & $15.9_{1.1}$ & $25.8_{0.4}$ & $17.5_{0.4}$ \\
& Temp   & $7.2_{2.6}$ & $9.4_{3.5}$ & $6.3_{1.3}$ & $16.4_{5.5}$ & $16.1_{4.6}$ & $22.7_{5.8}$ & $17.4_{6.0}$ \\
\cdashline{2-9} \rule{0pt}{2.25ex}
& LLLA   & $5.4_{0.4}$ & $6.5_{0.0}$ & $4.9_{0.0}$ & $\bm{12.6_{0.0}}$ & $\bm{12.6_{0.0}}$ & $\bm{15.7_{0.0}}$ & $\bm{15.3_{0.0}}$ \\
& LA     & $\bm{3.5_{0.4}}$ & $\bm{5.5_{0.0}}$ & $\bm{3.1_{0.0}}$ & $14.5_{0.0}$ & $12.8_{0.0}$ & $23.9_{0.0}$ & $17.6_{0.0}$ \\
\hline
\multirow{6}{*}{NLL $\downarrow$} \rule{0pt}{2.25ex}
& MAP    & $0.71_{0.03}$ & $0.95_{0.02}$ & $0.69_{0.02}$ & $1.40_{0.05}$ & $1.42_{0.07}$ & $1.81_{0.01}$ & $1.48_{0.02}$ \\
& MC Drop     & $0.71_{0.04}$ & $0.94_{0.02}$ & $0.68_{0.03}$ & $1.40_{0.05}$ & $1.41_{0.06}$ & $1.80_{0.01}$ & $1.47_{0.02}$ \\
& Ckpt Ens     & $0.65_{0.02}$ & $0.87_{0.00}$ & $\bm{0.61_{0.01}}$ & $1.29_{0.01}$ & $1.29_{0.03}$ & $1.71_{0.00}$ & $1.34_{0.01}$ \\
& Temp   & $0.67_{0.02}$ & $0.90_{0.05}$ & $0.66_{0.01}$ & $1.31_{0.06}$ & $1.32_{0.07}$ & $1.65_{0.16}$ & $1.36_{0.10}$ \\
\cdashline{2-9} \rule{0pt}{2.25ex}
& LLLA   & $0.66_{0.02}$ & $0.88_{0.00}$ & $0.64_{0.00}$ & $1.28_{0.00}$ & $\bm{1.27_{0.00}}$ & $\bm{1.49_{0.00}}$ & $\bm{1.31_{0.00}}$ \\
& LA     & $\bm{0.62_{0.01}}$ & $\bm{0.85_{0.00}}$ & $0.62_{0.00}$ & $\bm{1.26_{0.00}}$ & $\bm{1.27_{0.00}}$ & $1.68_{0.00}$ & $1.35_{0.00}$ \\
\end{tabular}}
\end{table}

\subsection{Fine-tuning RoBERTa for text classification} \label{app:roberta}
In this section, we present additional results of fine-tuning RoBERTa-base (Fig.~\ref{fig:roberta-base-kron}) and RoBERTa-large (Fig.~\ref{fig:roberta-large-kron}) \citep{liu2019roberta} with LoRA on text classification tasks from GLUE \citep{wang2018glue} and SuperGLUE \citep{wang2019superglue}. 
Results for RoBERTa-base are shown in Figure~\ref{fig:roberta-base-kron} and Table~\ref{table:roberta-base_train}, and results for RoBERTa-large are shown in Figure~\ref{fig:roberta-large-kron} and Table~\ref{table:roberta-large_train}.
Surprisingly, checkpoint ensemble and Monte-Carlo (MC) dropout exhibit distinct behavior on RoBERTa models compared to LlaMA2-7B. Checkpoint ensemble often performs much worse than the Maximum a Posteriori (MAP) estimation in terms of Expected Calibration Error (ECE) and Negative Log-Likelihood (NLL), while MC dropout often offers much more improvement on RoBERTa models. We suspect this difference arises due to an extra hidden penultimate layer with an additional dropout layer in front in the default RoBERTa fine-tuning setup; whereas, in LlaMA2-7B fine-tuning, we have LoRA weights and dropout on LoRA layers at the end. However, the gain from MC dropout diminishes on RoBERTa-large compared to RoBERTa-base.
On the other hand, Laplace-LoRA (LA) consistently delivers substantial gains on any models we have tested (RoBERTa-base, RoBERTa-large, and LlaMA2-7B), demonstrating the robustness of Laplace-LoRA. Moreover, the Last-Layer Laplace-LoRA (LLLA) offers modest improvements as in LlaMA2-7B when optimized by Laplace model evidence, underscoring the significance of performing Bayesian inference on all LoRA weights.


\begin{figure*}
    \centering
    \includegraphics[width=\textwidth]{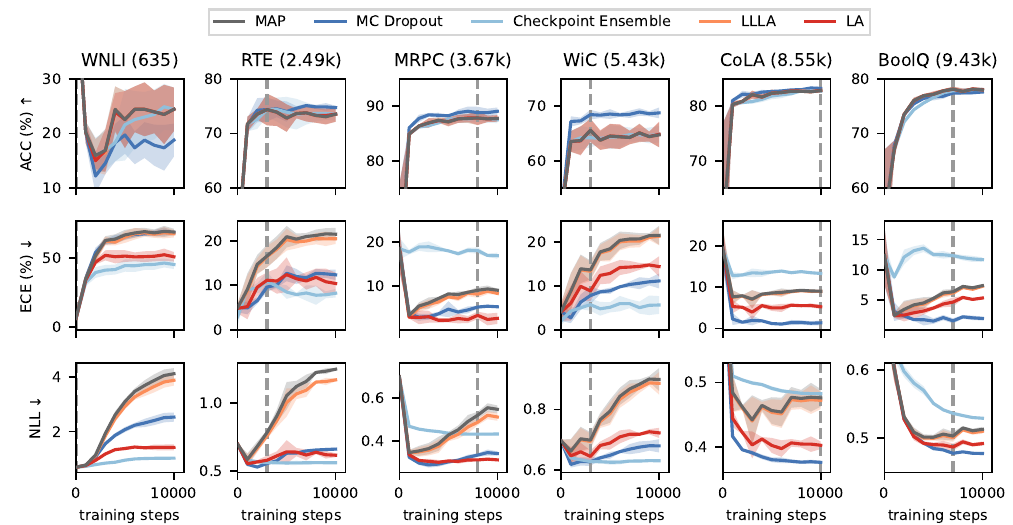}
    \caption{Fine-tuning of RoBERTa-base across six GLUE and SuperGLUE tasks (presented column-wise, with number of training examples in brackets), evaluated on the test set every 1000 gradient steps, without a validation set for hyperparameter tuning. The vertical dashed line gives the number of training steps with optimal MAP performance. Note that RoBERTa-base seems to fail on WNLI, but RoBERTa-large succeeds (Fig.~\ref{fig:roberta-large-kron}).
    \label{fig:roberta-base-kron}
    }
\end{figure*}

\begin{table}[t]
\small
\centering
\caption{Comparison of different post-hoc methods applied to the fine-tuned RoBERTa-base across six GLUE and SuperGLUE tasks, without a validation set for hyperparameter tuning. Results are evaluated at the early stopping point of 5000 gradient steps.}
\label{table:roberta-base_train}
\vspace{0.1cm}
\begin{tabular}{l|| l |l l l l l l}
Methods & Metrics & WNLI & RTE & MRPC & WiC & CoLA & BoolQ \\
\hline
\hline
\multirow{5}{*}{ACC $\uparrow$} \rule{0pt}{2.25ex}
& MAP    & $22.5_{4.6}$ & $72.8_{2.2}$ & $87.1_{0.7}$ & $64.5_{2.0}$ & $82.4_{0.6}$ & $77.2_{0.4}$ \\
& MC Drop  & $19.7_{3.0}$ & $74.0_{1.6}$ & $88.2_{0.0}$ & $68.5_{0.8}$ & $82.7_{0.1}$ & $76.7_{0.6}$ \\
& Ckpt Ens     & $21.6_{5.8}$ & $73.9_{2.5}$ & $87.7_{0.6}$ & $64.3_{1.9}$ & $81.8_{0.5}$ & $76.6_{0.2}$ \\
& LLLA   & $22.5_{4.6}$ & $72.8_{2.2}$ & $87.1_{0.7}$ & $64.5_{2.0}$ & $82.4_{0.6}$ & $77.2_{0.4}$ \\
\cdashline{2-8} \rule{0pt}{2.25ex}
& LA     & $22.5_{4.6}$ & $72.8_{2.2}$ & $87.2_{0.7}$ & $64.5_{2.0}$ & $82.4_{0.6}$ & $77.2_{0.5}$ \\
\hline
\multirow{5}{*}{ECE $\downarrow$} \rule{0pt}{2.25ex}
& MAP    & $66.7_{3.2}$ & $20.9_{2.1}$ & $8.3_{0.1}$ & $18.4_{2.2}$ & $8.6_{0.5}$ & $5.3_{0.2}$ \\
& MC Drop  & $65.6_{1.8}$ & $12.7_{1.6}$ & $4.8_{1.6}$ & $8.9_{0.6}$ & $\bm{1.2_{0.1}}$ & $\bm{1.5_{0.4}}$ \\
& Ckpt Ens     & $\bm{44.6_{3.9}}$ & $\bm{8.5_{1.3}}$ & $17.9_{0.1}$ & $\bm{5.2_{1.0}}$ & $13.6_{0.3}$ & $12.5_{0.3}$ \\
\cdashline{2-8} \rule{0pt}{2.25ex}
& LLLA   & $65.2_{3.7}$ & $20.3_{2.2}$ & $7.6_{0.3}$ & $18.1_{2.2}$ & $8.5_{0.4}$ & $5.1_{0.1}$ \\
& LA     & $51.4_{3.8}$ & $12.4_{3.4}$ & $\bm{2.3_{0.4}}$ & $12.9_{2.0}$ & $5.0_{0.6}$ & $3.7_{0.2}$ \\
\hline
\multirow{5}{*}{NLL $\downarrow$} \rule{0pt}{2.25ex}
& MAP    & $3.10_{0.09}$ & $1.05_{0.09}$ & $0.43_{0.01}$ & $0.79_{0.03}$ & $0.46_{0.02}$ & $0.50_{0.00}$ \\
& MC Drop  & $2.07_{0.08}$ & $0.63_{0.01}$ & $\bm{0.30_{0.01}}$ & $0.65_{0.01}$ & $\bm{0.38_{0.00}}$ & $\bm{0.49_{0.00}}$ \\
& Ckpt Ens     & $\bm{0.96_{0.02}}$ & $\bm{0.56_{0.01}}$ & $0.44_{0.00}$ & $\bm{0.63_{0.01}}$ & $0.49_{0.00}$ & $0.55_{0.00}$ \\
\cdashline{2-8} \rule{0pt}{2.25ex}
& LLLA   & $2.92_{0.08}$ & $1.00_{0.09}$ & $0.41_{0.01}$ & $0.79_{0.03}$ & $0.45_{0.02}$ & $0.50_{0.00}$ \\
& LA     & $1.38_{0.06}$ & $0.64_{0.08}$ & $\bm{0.30_{0.00}}$ & $0.69_{0.02}$ & $0.40_{0.01}$ & $\bm{0.49_{0.00}}$ \\
\end{tabular}
\end{table}

\begin{figure*}[t]
    \centering
    \includegraphics[width=\textwidth]{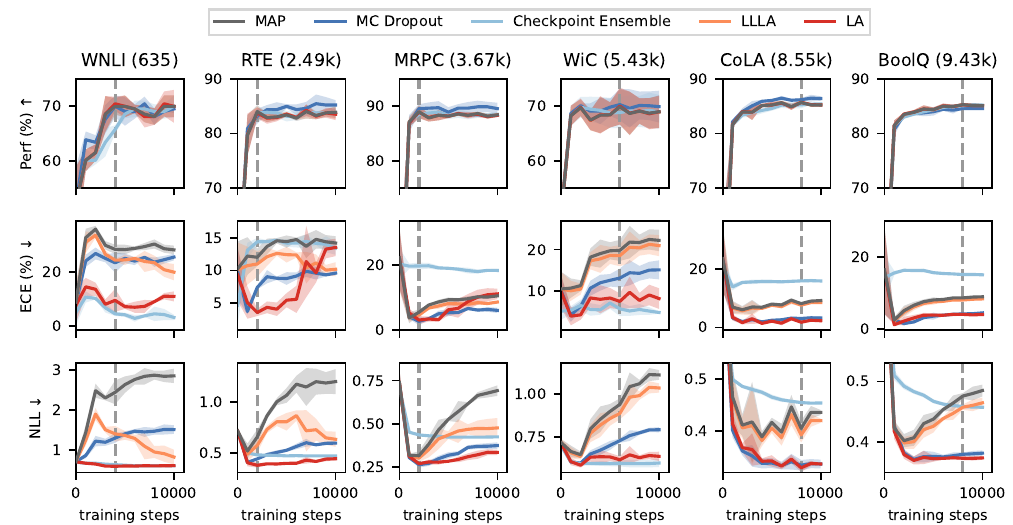}
    \caption{Fine-tuning RoBERTa-large across the six GLUE and SuperGLUE tasks in Fig.~\ref{fig:roberta-base-kron}, without a validation set for hyperparameter tuning. The vertical dashed line gives the number of training steps with optimal MAP performance.}
    \label{fig:roberta-large-kron}
\end{figure*}

\begin{table}[t]
\small
\centering
\caption{Comparison of different post-hoc methods applied to the fine-tuned RoBERTa-large across six GLUE and SuperGLUE tasks. Results are evaluated at the early stopping point of 5000 gradient steps.}
\label{table:roberta-large_train}
\vspace{0.1cm}
\begin{tabular}{l|| l |l l l l l l}
Methods & Metrics & WNLI & RTE & MRPC & WiC & CoLA & BoolQ \\
\hline
\hline
\multirow{5}{*}{ACC $\uparrow$} \rule{0pt}{2.25ex}
& MAP    & $70.0_{1.8}$ & $83.4_{0.3}$ & $88.2_{0.3}$ & $68.4_{2.0}$ & $85.2_{0.5}$ & $84.4_{0.4}$ \\
& MC Drop  & $69.0_{2.3}$ & $85.0_{0.3}$ & $88.9_{0.8}$ & $69.8_{1.4}$ & $86.0_{0.1}$ & $84.0_{0.6}$ \\
& Ckpt Ens     & $68.5_{0.7}$ & $83.2_{0.9}$ & $88.2_{0.5}$ & $68.8_{2.4}$ & $84.6_{0.4}$ & $84.3_{0.5}$ \\
\cdashline{2-8} \rule{0pt}{2.25ex}
& LLLA   & $70.0_{1.8}$ & $83.5_{0.3}$ & $88.2_{0.3}$ & $68.4_{2.0}$ & $85.2_{0.5}$ & $84.3_{0.3}$ \\
& LA     & $70.0_{1.8}$ & $83.5_{0.3}$ & $88.2_{0.3}$ & $68.4_{2.0}$ & $85.2_{0.5}$ & $84.4_{0.4}$ \\
\hline
\multirow{5}{*}{ECE $\downarrow$} \rule{0pt}{2.25ex}
& MAP    & $28.5_{1.6}$ & $14.4_{0.8}$ & $9.3_{0.6}$ & $19.7_{1.8}$ & $8.0_{0.4}$ & $8.0_{0.6}$ \\
& MC Drop  & $25.2_{1.6}$ & $8.8_{0.9}$ & $\bm{5.3_{1.0}}$ & $12.6_{1.3}$ & $2.4_{0.3}$ & $\bm{3.6_{0.9}}$ \\
& Ckpt Ens     & $\bm{4.2_{0.7}}$ & $14.2_{0.5}$ & $18.9_{0.2}$ & $\bm{7.2_{1.2}}$ & $15.7_{0.5}$ & $15.6_{0.3}$ \\
\cdashline{2-8} \rule{0pt}{2.25ex}
& LLLA   & $24.6_{1.7}$ & $12.4_{1.1}$ & $8.1_{0.6}$ & $18.6_{1.8}$ & $7.6_{0.3}$ & $7.5_{0.7}$ \\
& LA     & $7.2_{0.7}$ & $\bm{5.4_{1.0}}$ & $6.2_{1.5}$ & $8.4_{2.1}$ & $\bm{2.1_{0.3}}$ & $3.8_{0.5}$ \\
\hline
\multirow{5}{*}{NLL $\downarrow$} \rule{0pt}{2.25ex}
& MAP    & $2.64_{0.18}$ & $1.06_{0.06}$ & $0.53_{0.01}$ & $0.90_{0.04}$ & $0.42_{0.01}$ & $0.44_{0.01}$ \\
& MC Drop  & $1.40_{0.06}$ & $0.53_{0.01}$ & $0.32_{0.02}$ & $0.70_{0.02}$ & $\bm{0.34_{0.01}}$ & $0.38_{0.01}$ \\
& Ckpt Ens     & $0.62_{0.01}$ & $0.47_{0.00}$ & $0.42_{0.00}$ & $\bm{0.60_{0.01}}$ & $0.46_{0.00}$ & $0.47_{0.00}$ \\
\cdashline{2-8} \rule{0pt}{2.25ex}
& LLLA   & $1.36_{0.10}$ & $0.80_{0.08}$ & $0.44_{0.02}$ & $0.86_{0.04}$ & $0.41_{0.01}$ & $0.43_{0.01}$ \\
& LA     & $\bm{0.60_{0.02}}$ & $\bm{0.40_{0.01}}$ & $\bm{0.28_{0.00}}$ & $0.64_{0.02}$ & $\bm{0.34_{0.01}}$ & $\bm{0.37_{0.00}}$ \\
\end{tabular}
\end{table}

Similarly, we also conduct experiments by splitting the training set into a 80\% training set and a 20\% validation set, then tune temperature and Laplace prior precision on the validation set. The results are shown in Figure~\ref{fig:roberta-base-kron_val} and Table~\ref{table:roberta_base_id_val} for RoBERTa-base, and Figure~\ref{fig:roberta-large-kron_val} and Table~\ref{table:roberta_large_id_val} for RoBERTa-large. Again, full Laplace-LoRA LA offers the most improvements most of the time, and is the most robust method overall.

\begin{figure*}[t]
    \centering
    \includegraphics[width=\textwidth]{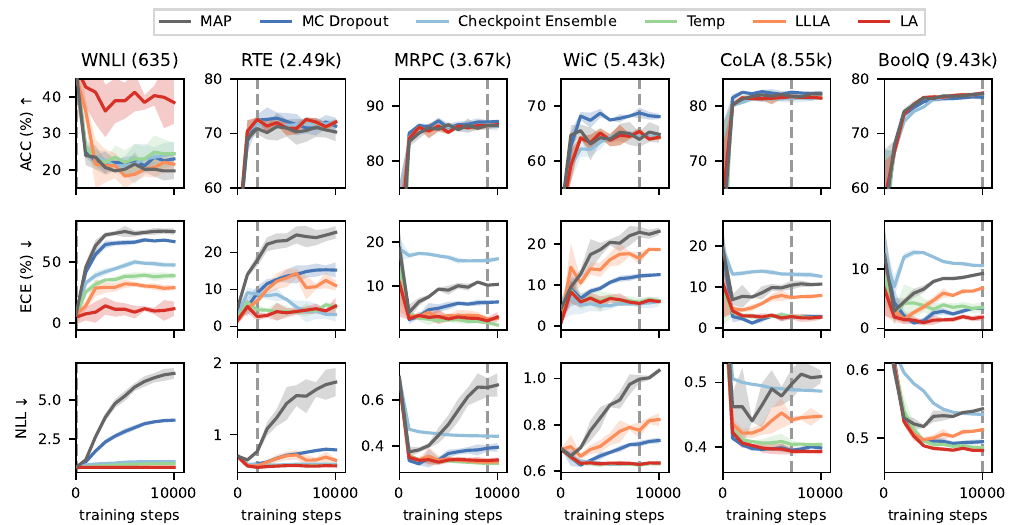}
    \caption{Fine-tuning RoBERTa-base across the six GLUE and SuperGLUE tasks, with a validation set for hyperparameter tuning. The vertical dashed line gives the checkpoint with optimal MAP performance on the validation set.}
    \label{fig:roberta-base-kron_val}
\end{figure*}

\begin{table}[t]
\small
\centering
\caption{Comparison of different post-hoc methods applied to the fine-tuned RoBERTa-base across six GLUE and SuperGLUE tasks, with a validation set for hyperparameter tuning. Results are evaluated at the best MAP performance checkpoint observed on the validation set.}
\label{table:roberta_base_id_val}
\vspace{0.1cm}
\begin{tabular}{l|| l |l l l l l l}
Methods & Metrics & WNLI & RTE & MRPC & WiC & CoLA & BoolQ \\
\hline
\hline
\multirow{6}{*}{ACC $\uparrow$} \rule{0pt}{2.25ex}
& MAP        & $47.9_{6.0}$ & $70.9_{0.9}$ & $86.4_{0.6}$ & $63.9_{1.2}$ & $81.8_{0.3}$ & $77.2_{0.2}$ \\
& MC Drop    & $47.9_{6.0}$ & $72.4_{0.9}$ & $87.1_{0.3}$ & $68.8_{0.9}$ & $82.6_{0.1}$ & $76.6_{0.4}$ \\
& Ckpt Ens   & $47.9_{6.0}$ & $71.8_{0.9}$ & $86.3_{0.5}$ & $64.7_{0.3}$ & $81.4_{0.7}$ & $77.2_{0.1}$ \\
& Temp       & $47.9_{6.0}$ & $72.6_{0.3}$ & $86.5_{0.3}$ & $65.4_{0.8}$ & $81.8_{0.8}$ & $77.3_{0.2}$ \\
\cdashline{2-8} \rule{0pt}{2.25ex}
& LLLA       & $46.0_{10.0}$ & $72.6_{0.3}$ & $86.4_{0.4}$ & $65.3_{0.7}$ & $81.8_{0.7}$ & $77.4_{0.2}$ \\
& LA         & $48.4_{7.7}$ & $72.6_{0.3}$ & $86.4_{0.3}$ & $65.4_{0.6}$ & $81.7_{0.7}$ & $77.4_{0.2}$ \\
\hline
\multirow{6}{*}{ECE $\downarrow$} \rule{0pt}{2.25ex}
& MAP        & $6.3_{4.3}$ & $17.8_{1.4}$ & $10.2_{0.2}$ & $22.8_{1.1}$ & $10.4_{0.9}$ & $9.2_{0.1}$ \\
& MC Drop    & $6.4_{4.4}$ & $9.0_{1.7}$ & $6.3_{0.4}$ & $12.2_{0.5}$ & $\bm{2.8_{0.3}}$ & $3.4_{0.1}$ \\
& Ckpt Ens   & $6.2_{2.3}$ & $8.8_{0.9}$ & $15.7_{0.6}$ & $5.5_{1.0}$ & $13.1_{0.5}$ & $10.6_{0.1}$ \\
& Temp       & $5.5_{1.7}$ & $4.5_{1.3}$ & $\bm{1.9_{0.3}}$ & $5.8_{0.9}$ & $2.9_{0.6}$ & $3.5_{0.7}$ \\
\cdashline{2-8} \rule{0pt}{2.25ex}
& LLLA       & $6.8_{7.8}$ & $6.9_{0.7}$ & $2.6_{0.4}$ & $16.4_{1.4}$ & $7.4_{0.3}$ & $6.8_{0.2}$ \\
& LA         & $\bm{4.7_{5.2}}$ & $\bm{2.7_{1.0}}$ & $2.2_{0.7}$ & $\bm{5.5_{0.5}}$ & $\bm{2.8_{0.6}}$ & $\bm{1.9_{0.5}}$ \\
\hline
\multirow{6}{*}{NLL $\downarrow$} \rule{0pt}{2.25ex}
& MAP        & $0.70_{0.01}$ & $0.76_{0.03}$ & $0.66_{0.04}$ & $1.00_{0.02}$ & $0.50_{0.02}$ & $0.54_{0.00}$ \\
& MC Drop    & $0.70_{0.01}$ & $0.58_{0.03}$ & $0.39_{0.01}$ & $0.72_{0.01}$ & $0.39_{0.00}$ & $0.50_{0.00}$ \\
& Ckpt Ens   & $\bm{0.69_{0.00}}$ & $0.57_{0.00}$ & $0.44_{0.00}$ & $0.63_{0.00}$ & $0.49_{0.00}$ & $0.53_{0.00}$ \\
& Temp       & $\bm{0.69_{0.00}}$ & $\bm{0.54_{0.00}}$ & $\bm{0.32_{0.00}}$ & $\bm{0.62_{0.01}}$ & $0.40_{0.01}$ & $0.49_{0.00}$ \\
\cdashline{2-8} \rule{0pt}{2.25ex}
& LLLA       & $\bm{0.69_{0.00}}$ & $0.56_{0.01}$ & $0.33_{0.00}$ & $0.78_{0.03}$ & $0.44_{0.00}$ & $0.51_{0.00}$ \\
& LA         & $\bm{0.69_{0.00}}$ & $\bm{0.54_{0.00}}$ & $0.34_{0.01}$ & $\bm{0.62_{0.01}}$ & $\bm{0.39_{0.00}}$ & $\bm{0.48_{0.00}}$ \\
\end{tabular}
\end{table}

\begin{figure*}[t]
    \centering
    \includegraphics[width=\textwidth]{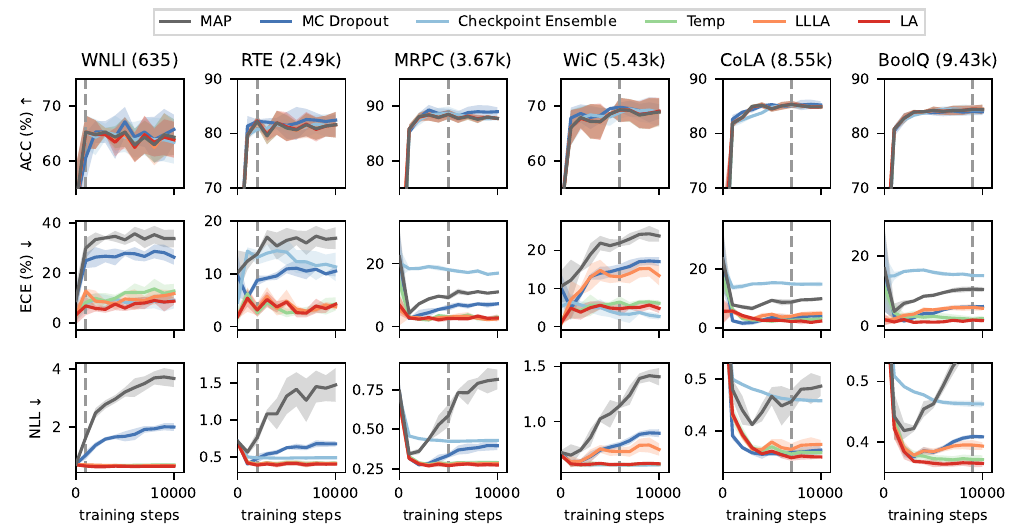}
    \caption{Fine-tuning RoBERTa-large across the six GLUE and SuperGLUE tasks, with a validation set for hyperparameter tuning. The vertical dashed line gives the checkpoint with optimal MAP performance on the validation set.}
    \label{fig:roberta-large-kron_val}
\end{figure*}

\begin{table}[t]
\small
\centering
\caption{Comparison of different post-hoc methods applied to the fine-tuned RoBERTa-large across six GLUE and SuperGLUE tasks, with a validation set split from the training set used for tuning temperature and Laplace prior precision. Results are evaluated at the best MAP performance checkpoint observed on the validation set.}
\label{table:roberta_large_id_val}
\vspace{0.1cm}
\begin{tabular}{l|| l |l l l l l l}
Methods & Metrics & WNLI & RTE & MRPC & WiC & CoLA & BoolQ \\
\hline
\hline
\multirow{6}{*}{ACC $\uparrow$} \rule{0pt}{2.25ex}
& MAP        & $65.3_{2.7}$ & $82.2_{0.9}$ & $88.5_{0.7}$ & $69.3_{1.9}$ & $85.3_{0.6}$ & $84.4_{0.6}$ \\
& MC Drop    & $60.6_{4.0}$ & $82.2_{0.3}$ & $88.7_{0.7}$ & $69.8_{0.8}$ & $85.1_{0.3}$ & $84.0_{0.6}$ \\
& Ckpt Ens   & $65.3_{2.7}$ & $80.7_{0.3}$ & $88.6_{1.1}$ & $68.8_{1.0}$ & $85.2_{0.4}$ & $84.3_{0.8}$ \\
& Temp       & $65.3_{2.7}$ & $82.2_{0.9}$ & $88.5_{0.7}$ & $69.3_{1.9}$ & $85.3_{0.6}$ & $84.4_{0.6}$ \\
\cdashline{2-8} \rule{0pt}{2.25ex}
& LLLA       & $65.3_{2.7}$ & $82.2_{0.9}$ & $88.5_{0.7}$ & $69.3_{1.9}$ & $85.3_{0.6}$ & $84.4_{0.6}$ \\
& LA         & $65.3_{2.7}$ & $82.3_{0.8}$ & $88.5_{0.7}$ & $69.3_{2.0}$ & $85.3_{0.5}$ & $84.4_{0.6}$ \\
\hline
\multirow{6}{*}{ECE $\downarrow$} \rule{0pt}{2.25ex}
& MAP        & $30.0_{3.5}$ & $13.7_{0.4}$ & $9.2_{0.7}$ & $22.0_{0.8}$ & $8.8_{0.8}$ & $10.5_{0.6}$ \\
& MC Drop    & $24.9_{2.6}$ & $8.8_{0.8}$ & $6.0_{0.9}$ & $15.1_{1.1}$ & $3.3_{0.8}$ & $5.5_{0.4}$ \\
& Ckpt Ens   & $7.7_{3.0}$ & $13.1_{1.2}$ & $18.0_{0.5}$ & $\bm{3.3_{1.6}}$ & $15.0_{0.3}$ & $14.4_{0.1}$ \\
& Temp       & $8.1_{2.2}$ & $\bm{2.6_{0.7}}$ & $3.1_{0.4}$ & $6.4_{1.3}$ & $2.7_{0.8}$ & $1.9_{0.7}$ \\
\cdashline{2-8} \rule{0pt}{2.25ex}
& LLLA       & $12.7_{0.9}$ & $3.3_{1.0}$ & $3.0_{0.5}$ & $13.1_{1.2}$ & $4.0_{0.4}$ & $5.3_{0.8}$ \\
& LA         & $\bm{6.8_{3.3}}$ & $3.1_{1.1}$ & $\bm{2.6_{0.7}}$ & $4.7_{1.7}$ & $\bm{2.2_{0.2}}$ & $\bm{1.7_{0.4}}$ \\
\hline
\multirow{6}{*}{NLL $\downarrow$} \rule{0pt}{2.25ex}
& MAP        & $1.63_{0.09}$ & $0.77_{0.07}$ & $0.59_{0.06}$ & $1.14_{0.02}$ & $0.46_{0.02}$ & $0.56_{0.01}$ \\
& MC Drop    & $1.02_{0.13}$ & $0.48_{0.06}$ & $0.33_{0.01}$ & $0.79_{0.02}$ & $0.36_{0.01}$ & $0.41_{0.00}$ \\
& Ckpt Ens   & $0.65_{0.02}$ & $0.49_{0.01}$ & $0.42_{0.00}$ & $\bm{0.61_{0.01}}$ & $0.46_{0.00}$ & $0.46_{0.00}$ \\
& Temp       & $0.65_{0.01}$ & $0.41_{0.01}$ & $0.28_{0.02}$ & $\bm{0.61_{0.01}}s$ & $0.36_{0.00}$ & $0.37_{0.01}$ \\
\cdashline{2-8} \rule{0pt}{2.25ex}
& LLLA       & $0.69_{0.02}$ & $0.40_{0.02}$ & $\bm{0.27_{0.01}}$ & $0.73_{0.03}$ & $0.36_{0.01}$ & $0.40_{0.01}$ \\
& LA         & $\bm{0.64_{0.00}}$ & $\bm{0.39_{0.01}}$ & $\bm{0.27_{0.01}}$ & $\bm{0.61_{0.01}}$ & $\bm{0.35_{0.01}}$ & $\bm{0.37_{0.01}}$ \\
\end{tabular}
\end{table}

\subsection{Fine-tuning LlaMA2-7B for common sense reasoning} \label{app:llama}
Figure~\ref{fig:llama7b-kron-val} displays the results of tuning temperature scaling and Laplace prior precision on a held-out validation set split from the training set. LLLA is missing a few zero checkpoint evaluation results due to Cholesky errors. Comparing Figure~\ref{fig:llama7b-kron-val} to Figure~\ref{fig:llama7b-kron} in the main text, it is evident that splitting the training set into a smaller training set (80\% data) and a validation set (20\% data) slightly impact the accuracy of fine-tuned model.

\begin{figure*}[t]
    \centering
    \includegraphics[width=\textwidth]{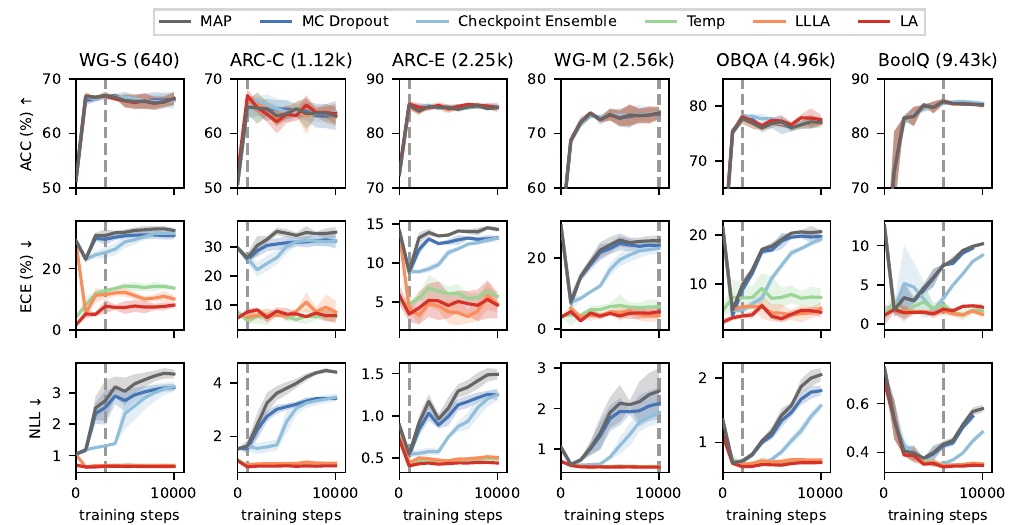}
    \caption{Fine-tuning of LlaMA2-7B across six common sense reasoning tasks (presented column-wise, with number of training examples in brackets), evaluated on the test set every 1000 gradient steps. The vertical dashed line gives the checkpoint with optimal MAP performance on a held-out validation set. }
    \label{fig:llama7b-kron-val}
\end{figure*}

\subsection{Fine-tuning Mistral-7B for common sense reasoning} \label{app:backbones}
In this section, we present the results using different decoder-based LLM backbones, specifically Mistral-7B \citep{jiang2023mistral}.
Table~\ref{table:mistral} shows the results of fine-tuning Mistral-7B on common-sense reasoning tasks. LA consistently offers the best NLL across all datasets, as well as 3 out of 6 best ECE, indicating the robustness of Laplace-LoRA on different decoder-only LLM backbones.

\begin{table}[t]
\small
\centering
\caption{Comparison of different post-hoc methods applied to the fine-tuned Mistral-7B across six common sense reasoning tasks. Results are evaluated at the early stopping point of 5000 gradient steps. We report standard deviations in subscripts, and bold numbers that are statistical significant.}
\label{table:mistral}
\vspace{0.1cm}
\begin{tabular}{l|| l |l l l l l l}
Methods & Metrics & WG-S & ARC-C & ARC-E & WG-M & OBQA & BoolQ \\
\hline
\hline
\multirow{5}{*}{ACC $\uparrow$} \rule{0pt}{2.25ex}
& MAP        & $77.4_{0.5}$ & $82.3_{1.0}$ & $90.7_{0.4}$ & $83.1_{0.8}$ & $87.0_{0.5}$ & $89.2_{0.1}$ \\
& MC Drop    & $77.4_{0.4}$ & $81.6_{1.4}$ & $90.7_{0.5}$ & $83.0_{1.0}$ & $87.1_{0.8}$ & $89.3_{0.1}$ \\
& Ckpt Ens   & $77.2_{0.2}$ & $81.8_{1.7}$ & $90.7_{0.7}$ & $83.8_{0.5}$ & $88.3_{0.5}$ & $89.0_{0.3}$ \\
\cdashline{2-8} \rule{0pt}{2.25ex}
& LLLA       & $77.4_{0.5}$ & $81.7_{1.0}$ & $90.6_{0.4}$ & $83.1_{0.8}$ & $87.0_{0.5}$ & $89.2_{0.1}$ \\
& LA         & $77.2_{0.3}$ & $79.9_{3.3}$ & $89.4_{0.9}$ & $83.0_{0.8}$ & $87.6_{0.8}$ & $89.2_{0.1}$ \\
\hline
\multirow{5}{*}{ECE $\downarrow$} \rule{0pt}{2.25ex}
& MAP        & $22.0_{0.6}$ & $16.1_{1.3}$ & $8.5_{0.6}$ & $13.5_{1.5}$ & $9.3_{0.0}$ & $5.6_{0.6}$ \\
& MC Drop    & $20.5_{0.3}$ & $15.2_{1.5}$ & $7.6_{0.4}$ & $13.0_{1.6}$ & $8.8_{0.5}$ & $5.3_{0.6}$ \\
& Ckpt Ens   & $21.1_{0.3}$ & $\bm{14.7_{2.3}}$ & $\bm{7.1_{0.8}}$ & $10.1_{0.6}$ & $\bm{6.3_{0.3}}$ & $2.0_{0.6}$ \\
\cdashline{2-8} \rule{0pt}{2.25ex}
& LLLA       & $10.2_{0.5}$ & $29.4_{3.1}$ & $19.1_{7.7}$ & $12.3_{1.4}$ & $7.6_{0.4}$ & $5.4_{0.5}$ \\
& LA         & $\bm{8.2_{0.3}}$ & $21.8_{6.7}$ & $21.6_{8.1}$ & $\bm{6.6_{1.7}}$ & $7.2_{0.9}$ & $\bm{1.8_{0.4}}$  \\
\hline
\multirow{5}{*}{NLL $\downarrow$} \rule{0pt}{2.25ex}
& MAP        & $2.38_{0.27}$ & $1.60_{0.16}$ & $0.75_{0.12}$ & $0.80_{0.14}$ & $0.57_{0.03}$ & $0.32_{0.02}$ \\
& MC Drop    & $2.12_{0.20}$ & $1.41_{0.10}$ & $0.62_{0.07}$ & $0.75_{0.12}$ & $0.54_{0.03}$ & $0.31_{0.02}$ \\
& Ckpt Ens   & $2.04_{0.13}$ & $1.33_{0.16}$ & $\bm{0.48_{0.02}}$ & $0.55_{0.02}$ & $0.45_{0.01}$ & $\bm{0.27_{0.01}}$ \\
\cdashline{2-8} \rule{0pt}{2.25ex}
& LLLA       & $0.54_{0.01}$ & $0.82_{0.01}$ & $\bm{0.48_{0.08}}$ & $0.67_{0.07}$ & $0.47_{0.05}$ & $0.32_{0.02}$ \\
& LA         & $\bm{0.52_{0.02}}$ & $\bm{0.80_{0.15}}$ & $\bm{0.48_{0.11}}$ & $\bm{0.42_{0.01}}$ & $\bm{0.37_{0.01}}$ & $\bm{0.27_{0.01}}$ \\
\end{tabular}
\end{table}

\subsection{Laplace-LoRA on different subsets of layers}
Besides applying Laplace approximation on all layers or only the last layer, we can also apply Laplace approximation on a subset of layers, for example, the first $k$ layers. In Table~\ref{table:laplace_topk}, we present the results of applying Laplace approximation to the top 8 layers, 16 layers, 24 layers, and all layers. Although there is no conclusive trend in the results, applying Laplace approximation to more than 8 layers usually leads to improved NLL.

\begin{table}[t]
\small
\centering
\caption{Laplace-LoRA applied to different subsets of layers on the fine-tuned LlaMA2-7B across six common sense reasoning tasks. Results are evaluated at the early stopping point of 5000 gradient steps. We report standard deviations in subscripts, and bold numbers that are statistical significant.}
\label{table:laplace_topk}
\vspace{0.1cm}
\begin{tabular}{l|| l |l l l l l l}
Methods & Metrics & WG-S & ARC-C & ARC-E & WG-M & OBQA & BoolQ \\
\hline
\hline
\multirow{4}{*}{ACC $\uparrow$} \rule{0pt}{2.25ex}
& LA 8   & $67.4_{0.8}$ & $65.6_{0.4}$ & $84.7_{1.5}$ & $73.3_{0.8}$ & $79.3_{0.4}$ & $85.9_{0.3}$ \\
& LA 16   & $67.6_{0.8}$ & $65.6_{0.4}$ & $84.6_{1.4}$ & $73.4_{0.7}$ & $79.3_{0.2}$ & $85.9_{0.3}$ \\
& LA 24   & $67.5_{0.8}$ & $65.1_{0.5}$ & $84.5_{1.4}$ & $73.4_{0.7}$ & $79.3_{0.2}$ & $85.9_{0.3}$ \\
& LA   & $67.3_{0.2}$ & $65.3_{0.2}$ & $85.1_{1.5}$ & $73.4_{0.3}$ & $78.9_{0.2}$ & $86.1_{0.2}$ \\
\hline
\multirow{4}{*}{ECE $\downarrow$} \rule{0pt}{2.25ex}
& LA 8   & ${6.4_{0.9}}$ & ${3.6_{0.4}}$ & ${4.3_{1.6}}$ & ${8.7_{0.6}}$ & ${6.3_{0.3}}$ & ${2.3_{0.5}}$ \\
& LA 16   & ${2.8_{1.0}}$ & ${5.8_{0.6}}$ & ${5.3_{0.3}}$ & ${5.8_{0.5}}$ & ${4.6_{0.7}}$ & ${1.4_{0.0}}$ \\
& LA 24   & ${2.0_{1.1}}$ & ${7.2_{1.1}}$ & ${6.6_{1.7}}$ & ${5.2_{0.5}}$ & ${3.7_{0.4}}$ & ${1.3_{0.0}}$ \\
& LA  & ${2.1_{0.3}}$ & ${7.4_{0.7}}$ & ${5.4_{0.2}}$ & ${7.4_{0.4}}$ & ${6.4_{0.8}}$ & ${2.1_{0.6}}$ \\
\hline
\multirow{4}{*}{NLL $\downarrow$} \rule{0pt}{2.25ex}
& LA 8   & ${0.63_{0.00}}$ & ${0.87_{0.03}}$ & ${0.50_{0.08}}$ & ${0.67_{0.01}}$ & ${0.68_{0.01}}$ & ${0.34_{0.00}}$ \\
& LA 16   & ${0.61_{0.01}}$ & ${0.86_{0.02}}$ & ${0.48_{0.06}}$ & ${0.59_{0.01}}$ & ${0.63_{0.01}}$ & ${0.33_{0.00}}$ \\
& LA 24   & ${0.60_{0.01}}$ & ${0.87_{0.02}}$ & ${0.48_{0.05}}$ & ${0.58_{0.01}}$ & ${0.62_{0.01}}$ & ${0.33_{0.00}}$ \\
& LA  & ${0.60_{0.01}}$ & ${0.88_{0.03}}$ & ${0.49_{0.06}}$ & ${0.63_{0.02}}$ & ${0.65_{0.01}}$ & ${0.34_{0.01}}$ \\
\end{tabular}
\end{table}

\subsection{Diagonal Laplace approximation} \label{app:diag}
In this section, we present the results for Laplace-LoRA utilizing a diagonal approximation of the Hessian. This approach is generally not found to be as effective as the Kronecker-factored Approximate Curvature (K-FAC) \citep{daxberger2021laplace} that approximates the block diagonal Hessian.

\subsubsection{RoBERTa}
The results on RoBERTa-base and RoBERTa-large are shown in Figure~\ref{fig:roberta-base-diag} Table~\ref{table:roberta-base_train_diag}, and Figure~\ref{fig:roberta-large-diag} Table~\ref{table:roberta-large_train_diag}. LLLA still offers a tiny advantage over Maximum a Posteriori (MAP) in ECE and NLL, however, LA show mixed performance across the datasets. 

\begin{figure*}[t]
    \centering
    \includegraphics[width=\textwidth]{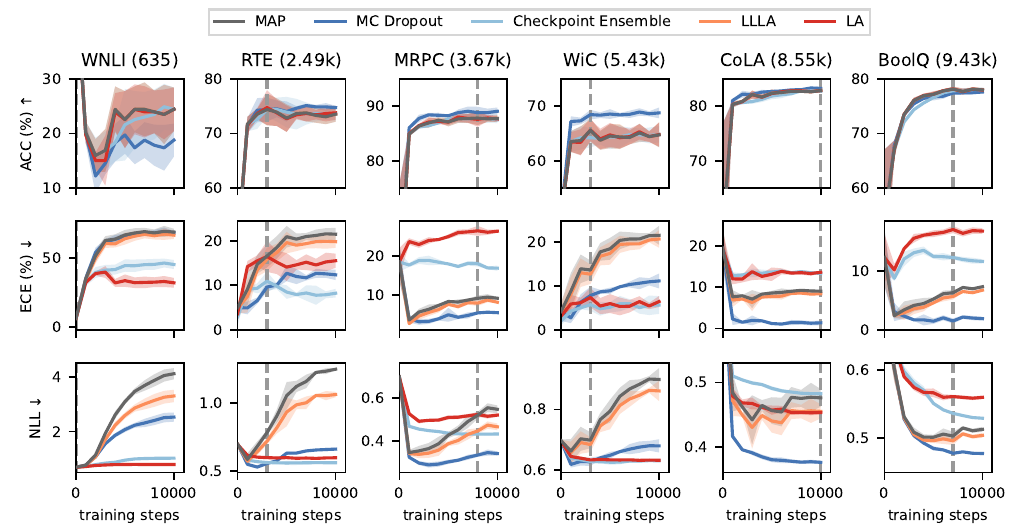}
    \caption{Fine-tuning of RoBERTa-base across six GLUE and SuperGLUE tasks (presented column-wise, with number of training examples in brackets), evaluated on the test set every 1000 gradient steps. The vertical dashed line gives the number of training steps with optimal MAP performance. LA and LLLA using diagonal Fisher approximation.}
    \label{fig:roberta-base-diag}
\end{figure*}

\begin{table}[t]
\small
\centering
\caption{Comparison of different post-hoc methods applied to the fine-tuned RoBERTa-base across six common GLUE and SuperGLUE tasks. Results are evaluated at the early stopping point of 5000 gradient steps. LA and LLLA using diagonal Fisher approximation.}
\label{table:roberta-base_train_diag}
\vspace{0.1cm}
\begin{tabular}{l|| l |l l l l l l}
Methods & Metrics & WNLI & RTE & MRPC & WiC & CoLA & BoolQ \\
\hline
\hline
\multirow{5}{*}{ACC $\uparrow$} \rule{0pt}{2.25ex}
& MAP    & $22.5_{4.6}$ & $72.8_{2.2}$ & $87.1_{0.7}$ & $64.5_{2.0}$ & $82.4_{0.6}$ & $77.2_{0.4}$ \\
& MC Drop  & $19.7_{3.0}$ & $74.0_{1.6}$ & $88.2_{0.0}$ & $68.5_{0.8}$ & $82.7_{0.1}$ & $76.7_{0.6}$ \\
& Ckpt Ens     & $21.6_{5.8}$ & $73.9_{2.5}$ & $87.7_{0.6}$ & $64.3_{1.9}$ & $81.8_{0.5}$ & $76.6_{0.2}$ \\
\cdashline{2-8} \rule{0pt}{2.25ex}
& LLLA   & $22.5_{4.6}$ & $72.8_{2.2}$ & $87.1_{0.7}$ & $64.5_{2.0}$ & $82.4_{0.6}$ & $77.2_{0.4}$ \\
& LA     & $22.5_{4.6}$ & $72.7_{2.1}$ & $87.1_{0.5}$ & $64.7_{1.9}$ & $82.3_{0.7}$ & $77.2_{0.5}$ \\
\hline
\multirow{5}{*}{ECE $\downarrow$} \rule{0pt}{2.25ex}
& MAP    & $66.7_{3.2}$ & $20.9_{2.1}$ & $8.3_{0.1}$ & $18.4_{2.2}$ & $8.6_{0.5}$ & $5.3_{0.2}$ \\
& MC Drop  & $65.6_{1.8}$ & $12.7_{1.6}$ & $\bm{4.8_{1.6}}$ & $8.9_{0.6}$ & $\bm{1.2_{0.1}}$ & $\bm{1.5_{0.4}}$ \\
& Ckpt Ens     & $44.6_{3.9}$ & $\bm{8.5_{1.3}}$ & $17.9_{0.1}$ & $\bm{5.2_{1.0}}$ & $13.6_{0.3}$ & $12.5_{0.3}$ \\
\cdashline{2-8} \rule{0pt}{2.25ex}
& LLLA   & $64.0_{3.7}$ & $19.6_{2.1}$ & $7.5_{0.2}$ & $17.4_{2.3}$ & $7.7_{0.5}$ & $4.6_{0.1}$ \\
& LA     & $\bm{33.7_{4.0}}$ & $14.1_{2.8}$ & $25.0_{0.5}$ & $6.0_{1.9}$ & $13.1_{1.4}$ & $16.4_{0.5}$ \\
\hline
\multirow{5}{*}{NLL $\downarrow$} \rule{0pt}{2.25ex}
& MAP    & $3.10_{0.09}$ & $1.05_{0.09}$ & $0.43_{0.01}$ & $0.79_{0.03}$ & $0.46_{0.02}$ & $0.50_{0.00}$ \\
& MC Drop  & $2.07_{0.08}$ & $0.63_{0.01}$ & $\bm{0.30_{0.01}}$ & $0.65_{0.01}$ & $\bm{0.38_{0.00}}$ & $\bm{0.49_{0.00}}$ \\
& Ckpt Ens     & $0.96_{0.02}$ & $\bm{0.56_{0.01}}$ & $0.44_{0.00}$ & $\bm{0.63_{0.01}}$ & $0.49_{0.00}$ & $0.55_{0.00}$ \\
\cdashline{2-8} \rule{0pt}{2.25ex}
& LLLA   & $2.59_{0.10}$ & $0.93_{0.09}$ & $0.39_{0.01}$ & $0.77_{0.03}$ & $0.44_{0.02}$ & $0.50_{0.00}$ \\
& LA     & $\bm{0.80_{0.00}}$ & $0.60_{0.01}$ & $0.51_{0.01}$ & $\bm{0.63_{0.00}}$ & $0.46_{0.01}$ & $0.57_{0.01}$ \\
\end{tabular}
\end{table}

\begin{figure*}[t]
    \centering
    \includegraphics[width=\textwidth]{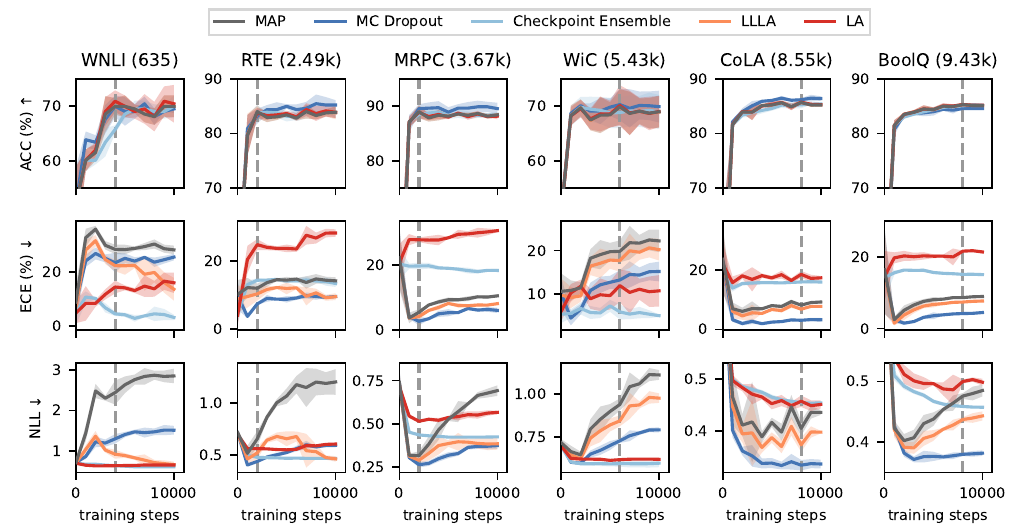}
    \caption{Fine-tuning of RoBERTa-large across six GLUE and SuperGLUE tasks (presented column-wise, with number of training examples in brackets), evaluated on the test set every 1000 gradient steps. The vertical dashed line gives the number of training steps with optimal MAP performance. LA and LLLA using diagonal Fisher approximation.}
    \label{fig:roberta-large-diag}
\end{figure*}

\begin{table}[t]
\small
\centering
\caption{Comparison of different post-hoc methods applied to the fine-tuned RoBERTa-large across six common GLUE and SuperGLUE tasks. Results are evaluated at the early stopping point of 5000 gradient steps. LA and LLLA using diagonal Fisher approximation.}
\label{table:roberta-large_train_diag}
\vspace{0.1cm}
\begin{tabular}{l|| l |l l l l l l}
Methods & Metrics & WNLI & RTE & MRPC & WiC & CoLA & BoolQ \\
\hline
\hline
\multirow{5}{*}{ACC $\uparrow$} \rule{0pt}{2.25ex}
& MAP        & $70.0_{1.8}$ & $83.4_{0.3}$ & $88.2_{0.3}$ & $68.4_{2.0}$ & $85.2_{0.5}$ & $84.4_{0.4}$ \\
& MC Drop    & $69.0_{2.3}$ & $85.0_{0.3}$ & $88.9_{0.8}$ & $69.8_{1.4}$ & $86.0_{0.1}$ & $84.0_{0.6}$ \\
& Ckpt Ens   & $68.5_{0.7}$ & $83.2_{0.9}$ & $88.2_{0.5}$ & $68.8_{2.4}$ & $84.6_{0.4}$ & $84.3_{0.5}$ \\
\cdashline{2-8} \rule{0pt}{2.25ex}
& LLLA       & $70.0_{1.8}$ & $83.5_{0.3}$ & $88.2_{0.3}$ & $68.4_{2.0}$ & $85.2_{0.5}$ & $84.4_{0.4}$ \\
& LA         & $70.0_{1.8}$ & $83.6_{0.5}$ & $88.0_{0.3}$ & $68.5_{2.2}$ & $85.3_{0.6}$ & $84.4_{0.4}$ \\
\hline
\multirow{5}{*}{ECE $\downarrow$} \rule{0pt}{2.25ex}
& MAP        & $28.5_{1.6}$ & $14.4_{0.8}$ & $9.3_{0.6}$ & $19.7_{1.8}$ & $8.0_{0.4}$ & $8.0_{0.6}$ \\
& MC Drop    & $25.2_{1.6}$ & $\bm{8.8_{0.9}}$ & $\bm{5.3_{1.0}}$ & $12.6_{1.3}$ & $\bm{2.4_{0.3}}$ & $\bm{3.6_{0.9}}$ \\
& Ckpt Ens   & $\bm{4.2_{0.7}}$ & $14.2_{0.5}$ & $18.9_{0.2}$ & $\bm{7.2_{1.2}}$ & $15.7_{0.5}$ & $15.6_{0.3}$ \\
\cdashline{2-8} \rule{0pt}{2.25ex}
& LLLA       & $22.4_{3.0}$ & $12.0_{1.5}$ & $8.2_{0.3}$ & $17.8_{1.8}$ & $6.7_{0.4}$ & $6.7_{0.7}$ \\
& LA         & $14.2_{0.8}$ & $23.8_{1.0}$ & $28.3_{0.2}$ & $9.5_{2.8}$ & $17.3_{0.6}$ & $20.2_{0.5}$ \\
\hline
\multirow{5}{*}{NLL $\downarrow$} \rule{0pt}{2.25ex}
& MAP        & $2.64_{0.18}$ & $1.06_{0.06}$ & $0.53_{0.01}$ & $0.90_{0.04}$ & $0.42_{0.01}$ & $0.44_{0.01}$ \\
& MC Drop    & $1.40_{0.06}$ & $0.53_{0.01}$ & $\bm{0.32_{0.02}}$ & $0.70_{0.02}$ & $\bm{0.34_{0.01}}$ & $\bm{0.38_{0.01}}$ \\
& Ckpt Ens   & $\bm{0.62_{0.01}}$ & $\bm{0.47_{0.00}}$ & $0.42_{0.00}$ & $\bm{0.60_{0.01}}$ & $0.46_{0.00}$ & $0.47_{0.00}$ \\
\cdashline{2-8} \rule{0pt}{2.25ex}
& LLLA       & $0.88_{0.05}$ & $0.66_{0.06}$ & $0.39_{0.02}$ & $0.82_{0.03}$ & $0.39_{0.01}$ & $0.41_{0.01}$ \\
& LA         & $0.64_{0.01}$ & $0.55_{0.01}$ & $0.54_{0.01}$ & $0.62_{0.00}$ & $0.45_{0.01}$ & $0.50_{0.00}$ \\
\end{tabular}
\end{table}


\subsubsection{LlaMA2-7B}
On LlaMA2-7B, both LLLA and LA show mixed performance and do not offer consistent improvements as shown in Figure~\ref{fig:llama7b-diag} and Table~\ref{table:id_train_diag}. Specifically, the accuracy of diagonal LA is much worse than MAP on ARC datasets, and ECE is worse than MAP on ARC-E, OBQA and BoolQ.

\begin{figure*}[t]
    \centering
    \includegraphics[width=\textwidth]{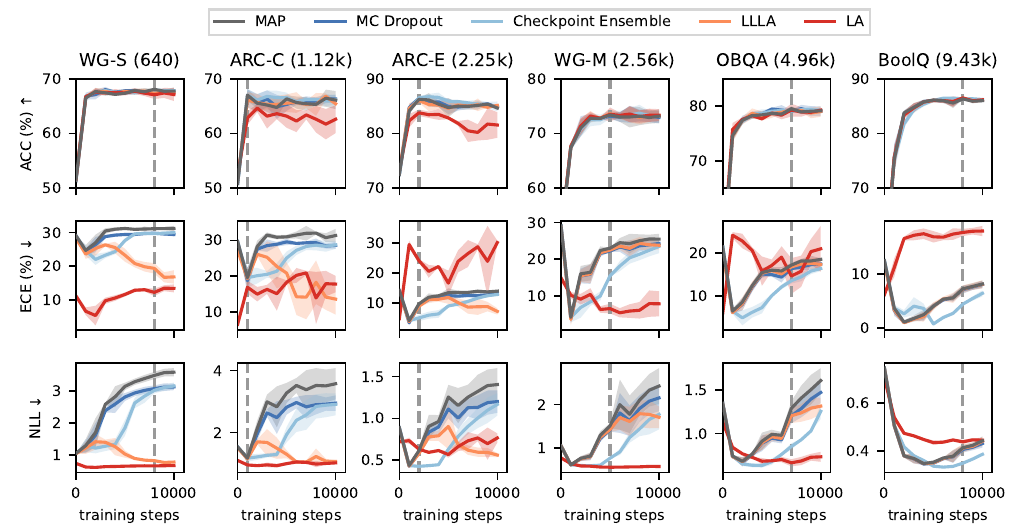}
    \caption{Fine-tuning of LlaMA2-7B across six common sense reasoning tasks (presented column-wise, with number of training examples in brackets), evaluated on the test set every 1000 gradient steps. Laplace (diagonal) prior precision is tuned using model evidence.}
    \label{fig:llama7b-diag}
\end{figure*}

\begin{table}[t]
\small
\centering
\caption{Comparison of different post-hoc methods applied to the fine-tuned LlaMA2-7B across six common sense reasoning tasks. Results are evaluated at the early stopping point of 5000 gradient steps.}
\label{table:id_train_diag}
\vspace{0.1cm}
\begin{tabular}{l|| l |l l l l l l}
Methods & Metrics & WG-S & ARC-C & ARC-E & WG-M & OBQA & BoolQ \\
\hline
\hline
\multirow{5}{*}{ACC $\uparrow$} \rule{0pt}{2.25ex}
& MAP    & $67.4_{0.3}$ & $66.3_{0.6}$ & $84.7_{1.5}$ & $73.4_{0.4}$ & $78.7_{0.4}$ & $86.1_{0.2}$ \\
& MC Drop  & $67.8_{0.1}$ & $65.3_{1.0}$ & $85.0_{1.3}$ & $73.2_{0.5}$ & $79.5_{0.2}$ & $86.0_{0.3}$ \\
& Ckpt Ens   & $67.4_{0.2}$ & $65.5_{0.4}$ & $85.8_{0.2}$ & $73.6_{0.7}$ & $79.1_{0.1}$ & $86.3_{0.2}$ \\
\cdashline{2-8} \rule{0pt}{2.25ex}
& LLLA   & $67.7_{0.3}$ & $65.5_{1.4}$ & $84.6_{1.2}$ & $73.6_{0.7}$ & $78.7_{0.7}$ & $86.0_{0.3}$ \\
& LA     & $67.4_{0.4}$ & $63.2_{1.9}$ & $82.8_{0.2}$ & $73.5_{1.1}$ & $78.7_{0.4}$ & $86.1_{0.3}$ \\
\hline
\multirow{5}{*}{ECE $\downarrow$} \rule{0pt}{2.25ex}
& MAP    & $31.2_{0.3}$ & $31.0_{0.5}$ & $13.4_{1.3}$ & $23.0_{0.1}$ & $16.1_{0.6}$ & $4.0_{0.5}$ \\
& MC Drop  & $29.4_{0.3}$ & $29.6_{0.8}$ & $12.4_{1.2}$ & $22.2_{0.2}$ & $15.0_{0.4}$ & $4.1_{0.4}$ \\
& Ckpt Ens   & $29.7_{0.6}$ & $27.0_{1.5}$ & $\bm{9.8_{0.6}}$ & $17.4_{0.9}$ & $\bm{12.4_{0.3}}$ & $\bm{1.2_{0.4}}$ \\
\cdashline{2-8} \rule{0pt}{2.25ex}
& LLLA   & $23.3_{1.7}$ & $20.9_{2.6}$ & $11.8_{1.9}$ & $22.5_{0.6}$ & $15.9_{0.3}$ & $4.1_{0.6}$ \\
& LA     & $\bm{11.6_{0.5}}$ & $\bm{18.3_{0.5}}$ & $16.6_{3.7}$ & $\bm{6.6_{1.2}}$ & $17.2_{1.2}$ & $17.0_{0.9}$ \\
\hline
\multirow{5}{*}{NLL $\downarrow$} \rule{0pt}{2.25ex}
& MAP    & $3.15_{0.10}$ & $3.28_{0.29}$ & $1.26_{0.13}$ & $1.51_{0.05}$ & $0.99_{0.05}$ & $0.35_{0.01}$ \\
& MC Drop  & $2.81_{0.11}$ & $2.82_{0.21}$ & $1.11_{0.10}$ & $1.41_{0.03}$ & $0.95_{0.04}$ & $0.35_{0.01}$ \\
& Ckpt Ens   & $2.58_{0.15}$ & $2.36_{0.34}$ & $0.80_{0.06}$ & $0.87_{0.06}$ & $0.76_{0.01}$ & $\bm{0.33_{0.00}}$ \\
\cdashline{2-8} \rule{0pt}{2.25ex}
& LLLA   & $1.02_{0.10}$ & $1.30_{0.11}$ & $0.90_{0.24}$ & $1.42_{0.05}$ & $0.96_{0.05}$ & $0.35_{0.01}$ \\
& LA     & $\bm{0.64_{0.01}}$ & $\bm{1.00_{0.04}}$ & $\bm{0.57_{0.03}}$ & $\bm{0.55_{0.01}}$ & $\bm{0.69_{0.01}}$ & $0.44_{0.01}$ \\
\end{tabular}
\end{table}

On the other hand, when we split a validation set from the training set and tune the Laplace prior precision using the validation log-likelihood, the results are much better, as shown in Figure~\ref{fig:llama7b-diag-val} and Table~\ref{table:id_val_diag}.

\begin{figure*}[t]
    \centering
    \includegraphics[width=\textwidth]{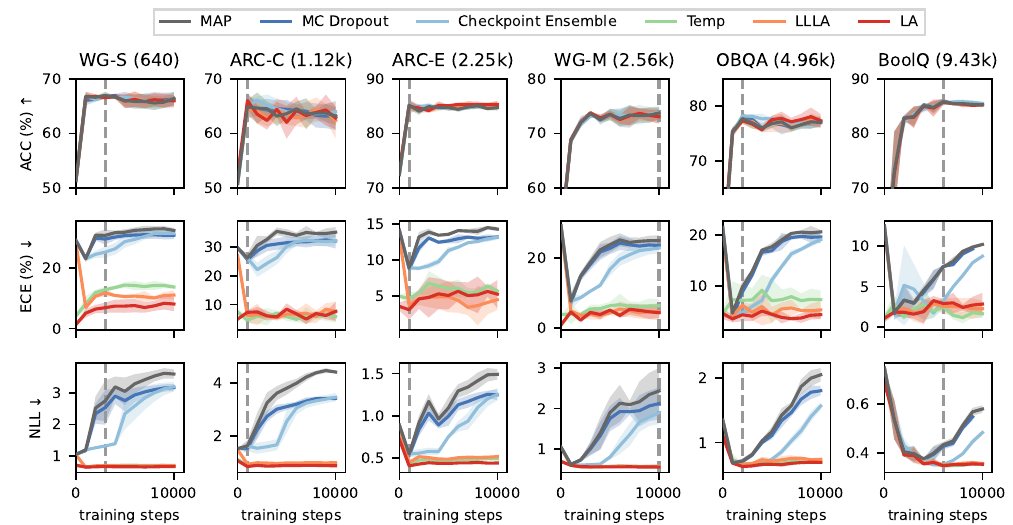}
    \caption{Fine-tuning of LlaMA2-7B across six common sense reasoning tasks (presented column-wise, with number of training examples in brackets), evaluated on the test set every 1000 gradient steps. The vertical dashed line gives the checkpoint with optimal MAP performance on a held-out validation set. Temperature scaling and Laplace (diagonal) prior precision are tuned on the validation set.}
    \label{fig:llama7b-diag-val}
\end{figure*}

\begin{table}[t]
\small
\centering
\caption{Comparison of different post-hoc methods applied to the fine-tuned LlaMA2-7B across six common sense reasoning tasks, with a validation set split from the training set used for tuning temperature and Laplace prior precision. Results are evaluated at the best MAP performance checkpoint observed on the validation set.}
\label{table:id_val_diag}
\vspace{0.1cm}
\begin{tabular}{l|| l |l l l l l l}
Methods & Metrics & WG-S & ARC-C & ARC-E & WG-M & OBQA & BoolQ \\
\hline
\hline
\multirow{6}{*}{ACC $\uparrow$} \rule{0pt}{2.25ex}
& MAP    & $67.0_{0.6}$ & $64.9_{1.1}$ & $85.2_{0.6}$ & $73.7_{0.9}$ & $77.7_{0.8}$ & $85.8_{0.4}$ \\
& MC Drop  & $66.7_{0.3}$ & $64.9_{1.9}$ & $85.1_{0.5}$ & $73.5_{0.9}$ & $77.7_{0.2}$ & $85.9_{0.4}$ \\
& Ckpt Ens     & $66.7_{0.3}$ & $64.9_{1.1}$ & $85.2_{0.6}$ & $73.8_{1.0}$ & $78.2_{0.2}$ & $85.4_{0.3}$ \\
& Temp   & $67.0_{0.6}$ & $64.9_{1.1}$ & $85.2_{0.6}$ & $73.7_{0.9}$ & $77.7_{0.8}$ & $85.8_{0.4}$ \\
\cdashline{2-8} \rule{0pt}{2.25ex}
& LLLA   & $66.7_{0.3}$ & $64.4_{1.0}$ & $85.1_{0.8}$ & $73.1_{1.2}$ & $77.2_{0.3}$ & $85.7_{0.4}$ \\
& LA     & $66.5_{0.3}$ & $66.0_{1.2}$ & $85.0_{1.2}$ & $73.1_{1.1}$ & $77.3_{0.3}$ & $85.7_{0.5}$ \\
\hline
\multirow{6}{*}{ECE $\downarrow$} \rule{0pt}{2.25ex}
& MAP    & $30.8_{1.8}$ & $26.1_{1.4}$ & $8.9_{0.3}$ & $24.9_{1.3}$ & $9.8_{1.0}$ & $7.4_{0.1}$ \\
& MC Drop  & $29.5_{1.6}$ & $25.6_{0.7}$ & $8.8_{0.6}$ & $23.5_{1.2}$ & $8.8_{0.8}$ & $7.5_{0.1}$ \\
& Ckpt Ens     & $25.2_{1.6}$ & $26.1_{1.4}$ & $8.9_{0.3}$ & $22.8_{1.4}$ & $4.7_{0.5}$ & $3.2_{0.5}$ \\
& Temp   & $12.8_{0.9}$ & $\bm{4.6_{1.0}}$ & $4.7_{0.8}$ & $6.3_{1.6}$ & $7.2_{2.6}$ & $\bm{2.5_{0.3}}$ \\
\cdashline{2-8} \rule{0pt}{2.25ex}
& LLLA   & $11.8_{1.1}$ & $6.0_{1.8}$ & $3.7_{0.5}$ & $4.5_{2.2}$ & $6.2_{0.5}$ & $2.6_{0.9}$ \\
& LA     & $\bm{6.9_{1.5}}$ & $7.3_{0.6}$ & $\bm{3.1_{1.2}}$ & $\bm{4.3_{1.3}}$ & $\bm{4.3_{1.1}}$ & $2.9_{0.5}$ \\
\hline
\multirow{6}{*}{NLL $\downarrow$} \rule{0pt}{2.25ex}
& MAP    & $2.75_{0.57}$ & $1.64_{0.19}$ & $0.54_{0.03}$ & $2.43_{0.50}$ & $0.71_{0.03}$ & $0.43_{0.01}$ \\
& MC Drop  & $2.54_{0.49}$ & $1.55_{0.16}$ & $0.52_{0.04}$ & $2.12_{0.35}$ & $0.71_{0.04}$ & $0.43_{0.01}$ \\
& Ckpt Ens     & $1.31_{0.04}$ & $1.64_{0.18}$ & $0.54_{0.03}$ & $1.89_{0.24}$ & $0.65_{0.02}$ & $\bm{0.35_{0.01}}$ \\
& Temp   & $0.68_{0.01}$ & $0.90_{0.01}$ & $0.43_{0.02}$ & $0.58_{0.01}$ & $0.67_{0.02}$ & $\bm{0.35_{0.00}}$ \\
\cdashline{2-8} \rule{0pt}{2.25ex}
& LLLA   & $0.68_{0.02}$ & $0.95_{0.03}$ & $0.44_{0.01}$ & $0.57_{0.01}$ & $0.66_{0.02}$ & $\bm{0.35_{0.00}}$ \\
& LA     & $\bm{0.66_{0.02}}$ & $\bm{0.86_{0.03}}$ & $\bm{0.40_{0.02}}$ & $\bm{0.55_{0.01}}$ & $\bm{0.63_{0.01}}$ & $\bm{0.35_{0.01}}$ \\
\end{tabular}
\end{table}

\section{LA vs LLLA, or which layers are uncertain?}

When comparing LA and LLLA, it is important to understand where the uncertainty arose from: the last layer (which is the only thing that LLLA captures), or the earlier layers (which is also captured by LA).
To understand this, we plotted the standard deviation of the logits arising from various sources (Fig.~\ref{fig:llama7b-kron-var-bar}).
We found that for full Laplace-LoRA LA, almost all the uncertainty arose from the lower-layers (green solid bar), rather than the last layer (blue solid bar).
Indeed, when optimizing the prior precisions using the model evidence, this implied that LLLA (orange solid bar) gave rise to far less uncertainty in the logits than LA.
This is likely a good explanation for the poor performance of LLLA when optimizing using the model evidence (Table~\ref{table:id_train}).
However, optimizing the prior precisions using the validation LL gives radically different results for LLLA, with much higher logit variances (dashed orange bar), and reasonable performance (Table~\ref{table:id_val}).
To understand what is going on here, consider a setting where alot of variability in the logits is necessary to optimize the validation LL. 
Then if we use the validation LL directly as an objective, LLLA can just increase the prior precision until the variance in the logits is high enough.
While this may be a reasonable strategy, the resulting validation-LL optimized LLLA posteriors differ dramatically from those we would expect under Bayes, where the uncertainty in at the last layer weights is far lower, and most of the uncertainty in logits arises at lower layers.

\begin{figure*}[t]
    \centering
    \includegraphics[width=\textwidth]{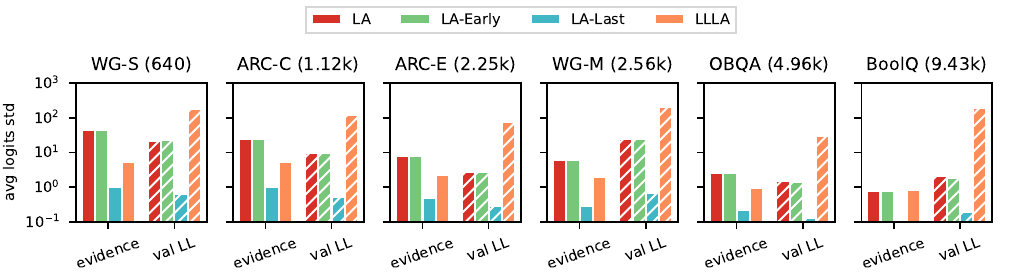}
    \caption{Bar chart of averaged logits standard deviation (bottom row) when optimizing Laplace model evidence (solid bars) on the training set at the early stopping checkpoint of 5000 training steps, and validation log-likelihood (dashed bars) on the validation set at the best MAP validation ACC checkpoint for each dataset. LA-Early involves taking a trained LA model, freezing the last layer and evaluating the uncertainty arising just from the ``earlier'' layers (i.e.\ all except the last layer).  Likewise, LA-Last involves taking a trained LA model, freezing the earlier layers (i.e.\ all except the last layer) and evaluating the uncertainty arising just from the last layer.  The ordering of the bars is the same as the ordering in the legend.}
    \label{fig:llama7b-kron-var-bar}
\end{figure*}

\section{Evaluation datasets under larger distribution shift} \label{app:mmlu}
Here we present the specific MMLU datasets we used for evaluations under distribution shift in Table~\ref{table:ood_train} and Table~\ref{table:ood_val}. The specific task splits we selected for each subject are shown in Table~\ref{table:mmlu} assigned by \citet{hendrycks2020measuring}.

\begin{table}[t]
\centering
\captionsetup{skip=8pt}
\begin{tabular}{c|c}
\hline
\textbf{Subject} & \textbf{Task} \\
\hline
Computer Science (CS) & college computer science \\ & computer security \\ & high school computer science \\ & machine learning \\
\hline
Engineering (Eng) & electrical engineering \\
\hline
Law & international law \\ & jurisprudence \\ & professional law \\
\hline
Health & anatomy \\ & clinical knowledge \\ & college medicine \\ & human aging \\ & nutrition \\ & professional medicine \\ & virology \\
\hline
\end{tabular}
\caption{MMLU subjects and tasks.}
\label{table:mmlu}
\end{table}

\section{Code implementations} \label{app:code}

We open sourced an original implementation based on Laplace Redux \citep{daxberger2021laplace} and ASDL \citep{osawa2023asdl} at \url{https://github.com/adamxyang/laplace-lora}, and a newer standalone implementation which we intended to support going forward at \url{https://github.com/MaximeRobeyns/bayesian_lora}.

\end{document}